%% file: main.tex
\documentclass[10pt,journal,compsoc]{IEEEtran}
\usepackage{amsmath,amsfonts}
\usepackage{algorithmic}
\usepackage{algorithm}
\usepackage{array} 
\usepackage[caption=false,font=normalsize,labelfont=sf,textfont=sf]{subfig}
\usepackage{textcomp}
\usepackage{stfloats}
\usepackage{url}
\usepackage{verbatim}
\usepackage{graphicx} 
\usepackage{hyperref}
\usepackage{cite}
\usepackage{xcolor}
\usepackage{amssymb}
\usepackage{booktabs}
\usepackage{color}
\usepackage{colortbl}
\usepackage{textcomp}
\usepackage{stfloats}
\usepackage{url}
\usepackage{verbatim}
\usepackage{graphicx}
\usepackage{cite}
\usepackage{marvosym}
\usepackage{booktabs} 
\usepackage{multirow}
\usepackage{tabularx}
\usepackage{ulem}

\usepackage{hyperref}
\usepackage{url}
\usepackage[utf8]{inputenc} 
\usepackage[T1]{fontenc}    
\usepackage{hyperref}       
\usepackage{url}            
\usepackage{booktabs}       
\usepackage{amsfonts}       
\usepackage{nicefrac}       
\usepackage{microtype}      
\usepackage{xcolor}         
\usepackage{amsmath}
\usepackage{amssymb}
\usepackage{mathtools}
\usepackage{amsthm}
\usepackage{epsfig}
\usepackage{amsfonts}
\usepackage{booktabs}
\usepackage{multirow}
\usepackage{float}
\usepackage{amsmath}
\usepackage{amssymb}
\usepackage{bm}
\usepackage{esint}
\usepackage{colortbl}
\usepackage{wrapfig}
\usepackage{bbm}
\usepackage{multirow}
\definecolor{origin}{rgb}{0.2,0.3,0.9}
\definecolor{mygray}{gray}{.9}
\definecolor{mygray2}{gray}{.8}
\usepackage{hyperref}
\usepackage{url}
\usepackage{graphicx}
\usepackage{amsmath}
\usepackage{mathrsfs}
\usepackage{wrapfig}
\usepackage{booktabs}       
\usepackage{amsfonts}       
\usepackage{nicefrac}       
\usepackage{microtype}      
\usepackage{multirow}
\usepackage{multicol}
\usepackage{makecell}
\usepackage{mathtools}
\usepackage{xspace}
\usepackage{amssymb}
\usepackage{algorithm}
\usepackage{algorithmic}
\usepackage{colortbl}
\usepackage{arydshln}
\usepackage{float}
\usepackage[dvipsnames]{xcolor}
\usepackage{makecell}
\usepackage{xspace}
\usepackage{pifont}



\makeatletter
\DeclareRobustCommand\onedot{\futurelet\@let@token\@onedot}
\def\@onedot{\ifx\@let@token.\else.\null\fi\xspace}

\def\eg{\emph{e.g}\onedot}

\makeatother

\DeclareRobustCommand{\ours}{VFMTok\xspace}

\makeatletter
\newcommand{\smaller}{\@setfontsize\mynotesize{8.4pt}{9.4pt}}
\makeatother

\newcommand{\tablestyle}[2]{\setlength{\tabcolsep}{#1}\renewcommand{\arraystretch}{#2}\centering\small}

\usepackage[capitalize]{cleveref}
\crefname{section}{Sec.}{Secs.}
\Crefname{section}{Section}{Sections}
\Crefname{table}{Table}{Tables}
\crefname{table}{Tab.}{Tabs.}
\Crefname{appendix}{Appendix}{Appendices}
\crefname{appendix}{Appx.}{Appxs.}

\definecolor{citecolor}{HTML}{0071BC}
\definecolor{linkcolor}{HTML}{ED1C24}
\definecolor{grey}{HTML}{999999}
\definecolor{green}{HTML}{ABD1BC}
\definecolor{lightblue}{HTML}{B0C4DE}
\definecolor{purple}{HTML}{E3BBED}
\definecolor{orange}{HTML}{ffdab9}
\newlength\savewidth

\theoremstyle{plain}
\definecolor{grass-green}{rgb}{0.4, 0.75, 0.4}

\theoremstyle{definition}

\theoremstyle{remark}

\definecolor{lightmauve}{rgb}{0.86, 0.82, 1.0}

\begin{document}

\title{Vision Foundation Models as Generalist Tokenizers for Image Generation}

\author{~~~ \\
        Anlin Zheng,~~~
        Qi Han,~~~
        Xin Wen~\IEEEmembership{Student Member, IEEE},~~~ 
        Chuofan Ma,~~~ \\
        Lanxi Gong,~~~ 
        Gang Yu,~~~ Xiangyu Zhang,~~~
        $\text{Xiaojuan Qi}^{\text{\Letter}}$~\IEEEmembership{Senior Member,~IEEE}
\IEEEcompsocitemizethanks{
\IEEEcompsocthanksitem Anlin Zheng, Xin Wen, Chuofan Ma, Lanxi Gong and Xiaojuan Qi are with the University of Hong Kong, Pokfulam, Hong Kong.  
\IEEEcompsocthanksitem Qi Han, Gang Yu, and Xiangyu Zhang are affiliated with StepFun.
\IEEEcompsocthanksitem{${\text{\Letter}}$ Corresponding to Xiaojuan Qi: \href{mailto:xjqi@eee.hku.hk}{xjqi@eee.hku.hk}.
\IEEEcompsocthanksitem A preliminary version of this research has published in NeurIPS 2025~\cite{vfm-tok}.}
}}

\newcommand{\anlin}[1]{{\color{red}{{[\textbf{anlin}: #1]}}}}


\maketitle

\input{sections/0_abstract}
\input{sections/1_introduction}
\input{sections/2_related_work}
\input{sections/3_approach}
\input{sections/4_experiments}

\input{sections/5_conclusion}

\bibliographystyle{plain}
\bibliography{main.bib}

\clearpage
\input{sections/6_appendix}

\vfill

\end{document}

%% file: sections/0_abstract.tex
\begin{abstract}

In this work, we explore the largely unexplored direction of building a generalist image tokenizer directly on top of a frozen vision foundation model (VFM). To build this tokenizer, we utilize a frozen VFM as the encoder and introduce two key innovations: (1) a region-adaptive quantization framework to eliminate spatial redundancy in standard 2D grid features, and (2) a semantic reconstruction objective that aligns the decoded outputs with the VFM's representations to preserve semantic fidelity.
Grounded in these designs, we propose \textbf{\ours}, a generalist visual tokenizer capable of operating seamlessly in both discrete and continuous latent spaces. \ours achieves substantial improvements in synthesis quality while drastically enhancing token efficiency. For discrete autoregressive (AR) generation, it accelerates model convergence by \textbf{3 times} and achieves a state-of-the-art gFID of \textbf{1.36} on ImageNet class-conditional synthesis. Similarly, for continuous-space generation, integrating \ours with a denoising model yields an exceptional gFID of \textbf{1.25}. Furthermore, because the latent space inherently captures rich spatial semantics, \ours enables high-fidelity class-conditional synthesis without classifier-free guidance (\textbf{w/o CFG}) across both generative paradigms, significantly accelerating inference speed.
Beyond these remarkable empirical results, we systematically investigate the underlying mechanisms of our approach. We discover that the specific self-supervised learning objectives utilized during VFM pre-training dictate its effectiveness as a tokenizer. Specifically, a VFM jointly optimized with global contrastive learning and latent masked image modeling provides the optimal representations for image tokenization. These insights establish a strong foundation and offer valuable guidance for the design of future image tokenizers.
The code is available at \href{https://github.com/CVMI-Lab/VFMTok}{https://github.com/CVMI-Lab/VFMTok}.

\end{abstract}

%% file: sections/1_introduction.tex
\vspace{-0.3cm}
\section{Introduction}

Recent advancements in image generation are largely driven by two dominant paradigms: autoregressive (AR) models~\cite{var, llamagen, hita, show-o} operating in discrete spaces and iterative denoising models~\cite{ddpm, nichol2021iddpm, liu2022flow, albergo2022building} operating in continuous spaces. Despite their distinct formulations, both paradigms rely on generating compressed latent representations rather than operating directly on raw pixels. This compression is achieved through an image tokenizer—a critical interface that maps raw pixels into a compact discrete or continuous space. Consequently, the tokenizer's design profoundly impacts synthesis quality, sampling efficiency, and the preservation of fine details.

\begin{figure*}[t]
  \centering
  \includegraphics[width=\textwidth]{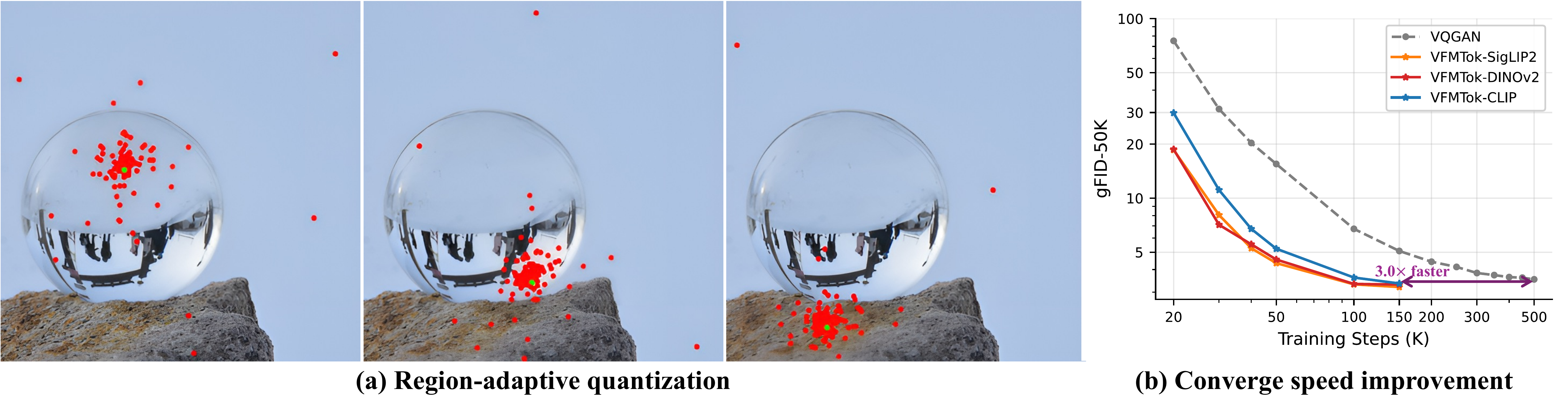}
  \caption{\ours introduces novel features, including: a).\textbf{region-adaptive quantization}— where it adaptively samples regions of similar patterns and extracts their VFM features for quantization; b).\textbf{convergence speed improvement} compared with vanilla VQGAN~\cite{llamagen} for AR image synthesis.}
  \label{fig:pilot2}
  \vspace{-.1cm}
\end{figure*}

Traditionally, AR models utilize discrete tokenizers like VQGAN~\cite{vqvae, vqgan, vit-vqgan, llamagen}, while denoising models adopt continuous autoencoders~\cite{ae, vincent2010stacked, diederik2019introduction} or VAEs~\cite{kingma2013auto, beta-vae}. Because these tokenizers are typically trained from scratch with a primary focus on pixel-level reconstruction, their latent spaces tend to be burdened with spatial redundancy and lack high-level semantics. These semantically deficient representations not only prolong the training convergence of AR models (\cref{fig:pilot2}), but also force generative models to rely on computationally expensive techniques like classifier-free guidance (CFG) for high-fidelity, class-conditional synthesis, thereby increasing inference latency.

Concurrently, pre-trained vision foundation models (VFMs)~\cite{ibot, dino, dinov2, dinov2reg, clip, siglip, siglip2} such as DINOv2~\cite{dinov2reg} and CLIP~\cite{clip} have demonstrated exceptional capabilities in extracting semantically rich and generalizable features. Early explorations (e.g., REPA~\cite{repa}) suggest that these semantic representations can facilitate the training of generative models. This motivates a compelling question: \textit{Can the representations derived from VFMs, originally designed for visual understanding, be repurposed as structured, foundational latents for both image reconstruction and generation?}

Recent studies have started exploring this direction by utilizing VFM features to initialize codebooks~\cite{vqgan-lc, zhu2024stabilize}, augment VQGAN architectures with additional branches~\cite{tokenflow}, or act as distillation targets for latent space learning~\cite{va-vae, repae, gigatok, dualtoken}. However, these approaches primarily treat VFM features as auxiliary guidance rather than intrinsic generative priors. Consequently, they suffer from architectural inefficiencies and fail to fully exploit the rich semantic information embedded within VFMs, leaving their potential as standalone, highly efficient tokenizers largely underexplored.

\textbf{Can VFMs be effective tokenizers?} 
To address this, we initialized the encoder of a VQGAN with different frozen pre-trained foundation models to reconstruct images. Once trained, the tokenizer is integrated on top of an AR model for image synthesis (implementation details depicted in \cref{method:pilot}) As shown in \cref{tbl:prelimiary_exp} (middle rows), our results demonstrate that \textbf{the semantically rich features from these foundation models not only support effective image reconstruction but also achieve generative performance comparable to---or even surpassing---that of a fully trained VQGAN encoder optimized for both reconstruction and generation}. These findings highlight the strong potential of pre-trained vision foundation models to serve as efficient and effective tokenizers for image generation tasks, eliminating the need for extensive encoder training while improving qualities.  

\textbf{Can we improve token efficiency for VFMs?} Building on this pilot study, we are further motivated by the observation that natural images often consist of irregular regions that exhibit recurring visual patterns. For example, as illustrated in \cref{fig:pilot2}(a), the upper portion of the crystal ball exhibits consistent patterns such as texture and transparency; similarly, the moss in the stone possesses similar textural structure. When such images are represented using a regular 2D feature grid extracted from foundation models, this structure-agnostic representation may introduce significant redundancy. Exploiting redundancy within semantically coherent regions offers a promising direction for improving tokenization efficiency. Motivated by this insight, we propose a region-adaptive strategy to refine the latent space that aims to enhance both image reconstruction and generation quality while significantly improving token representation efficiency.

\input{tables/pilot_study}

\textbf{Our solution and results.} Motivated by the preceding analysis, we introduce \ours, a generalist visual tokenizer that adopts a frozen vision foundation model (VFM) for efficient, adaptive region-level tokenization. Specifically, \ours utilizes a frozen VFM to extract multi-level semantic features. Learnable anchor queries, then perform region-level sampling on these features through deformable attention~\cite{deform_detr}, yielding region-adaptive representations. To ensure versatility, these features can be flexibly configured: they are either quantized into discrete codes for autoregressive (AR) modeling or compressed into low-dimensional continuous features for denoising synthesis. Subsequently, these representations are processed by a lightweight ViT~\cite{dosovitskiy2020image} in a BERT-style framework~\cite{bert, mae} that optimized by two primary objectives. By simultaneously reconstructing both the original image pixels and the VFM semantic features, \ours preserves the semantic richness and inherent discriminative power of the vision foundation model.

Once trained, \ours seamlessly bridges both autoregressive and denoising generative paradigms. The resulting compact, semantically rich latent features generated by either AR models~\cite{touvron2023llama, llamagen} or denoising models~\cite{ddpm, flow-matching} can be efficiently rendered into pixels. As highlighted in \cref{tbl:prelimiary_exp}, \ours achieves superior reconstruction and generation performance compared to its counterparts.

Extensive experiments validate that coupling VFM representations with our region-adaptive sampling strategy unlocks state-of-the-art image synthesis. First, \ours yields a highly compact, semantic-aware latent space, outperforming other tokenizers in both reconstruction quality and linear probing accuracy using only 256 tokens. Second, it significantly enhances downstream generation task. In discrete spaces, a 1.4B AR model equipped with \ours converges faster (\cref{fig:pilot2}b) and surpasses the LlamaGen-3B~\cite{llamagen} despite using fewer parameters; furthermore, our advanced 1.5B AR generator~\cite{rar} establishes a new state-of-the-art with a gFID of \textbf{1.36} on ImageNet. In continuous spaces, pairing \ours with RAE\cite{rae} yields a gFID of \textbf{1.25}, outperforming previous VFM-enhanced counterparts\cite{va-vae,repae, fae}. Finally, the compactness of \ours drastically accelerates AR inference. Crucially, the rich semantics embedded in the latents enable high-fidelity class-conditional synthesis without requiring classifier-free guidance (CFG), further minimizing inference latency across both generative paradigms.

\textbf{What makes a VFM a good image tokenizer?} 
To investigate how different VFM pre-training objectives influence tokenizer performance across reconstruction, generation, and semantic understanding, we conduct a comprehensive empirical study. Specifically, we evaluate VFMs optimized under distinct paradigms—namely, Contrastive Learning (C.L.)~\cite{clip}, pixel-level Masked Image Modeling (Pixel-MIM)~\cite{mae}, and latent-level MIM (Latent-MIM)~\cite{ibot}. Our findings indicate that a VFM jointly optimized with both Contrastive Learning and Latent-MIM strikes the optimal balance, thereby emerging as the most effective foundational backbone for visual tokenization and image synthesis.

\textbf{Difference from our conference paper.} This manuscript presents substantial extensions to our previous conference version~\cite{vfm-tok} in the following aspects:
(i) We generalize the discrete \ours into a continuous-valued AutoEncoder, tailored for continuous-valued image tokenization and synthesis. This substantiates that \ours can effectively serve as generalist visual tokenizers across both discrete and continuous latent spaces.
(ii) We reveal that for continuous-valued image generation, our \ours produces a compact, low-dimensional latent space that is highly compatible with denoising generative models~\cite{flow-matching,liu2022flow}. Consequently, it achieves comparable or even superior reconstruction and generation fidelity compared to existing counterparts~\cite{fae,rae,vfm-vae}.
(iii) We conduct extensive empirical evaluations to validate the robustness of our framework across both discrete and continuous settings, thereby strongly establishing frozen VFMs as powerful foundational components for efficient and high-quality image reconstruction and synthesis. 
(iv) We thoroughly investigate how different VFM pre-training objectives affect image reconstruction and generation. Our analysis reveals that the specific combination of these learning objectives is a critical factor to those tasks. Notably, a VFM jointly trained with Contrastive Learning (C.L.) and latent Masked Image Modeling (Latent-MIM) proves to be the optimal backbone for image tokenizers construction.

\noindent Our main contributions are summarized as follows: 
\begin{itemize}

    \item We demonstrate that frozen VFMs---whether self-supervised or language-supervised---serve as highly effective tokenizers. Their inherent semantic richness not only accelerates the convergence of autoregressive (AR) models but also enables both AR and denoising models to natively achieve high-fidelity, CFG-free image synthesis.

    \item We propose a novel region-adaptive quantization framework that exploits spatial redundancy to achieve highly compact tokenization. This strategy drastically reduces the sequence length of visual tokens, boosting AR inference efficiency without sacrificing generation quality.

    \item Through extensive experiments, we validate the effectiveness of \ours across both discrete and continuous-valued spaces, strongly establishing frozen VFMs as powerful foundational components for high-quality and efficient visual generation.

    \item We present a comprehensive empirical study analyzing the impact of different VFM pre-training objectives on tokenizer performance. The findings reveal that coupling Contrastive Learning with Latent Masked Image Modeling yields the optimal representations for image tokenization.

\end{itemize}

%% file: tables/pilot_study.tex
\begin{table}
    \caption{Pilot study of image reconstruction and generation on ImageNet~\cite{imagenet}. Relative wall-clock inference time for the tokenizer (compared to VFMTok) is reported. L.P. denotes linear probing results on the ImageNet validation set, used to estimate the semantic quality of latent tokens.}
	\label{tbl:prelimiary_exp}
    \vspace{-.35cm}
	\centering
    \scalebox{0.96}{
    \tablestyle{2.05pt}{1.2}
	\begin{tabular}{l|ccc|ccc|c}
        \Xhline{0.8pt}
    \multirow{2}{*}{Setup} & \multicolumn{3}{c|}{Image Recon.} & \multicolumn{3}{c|}{AR Generation} & L.P. \\
    & \#Tok. & rFID$\downarrow$ & rIS$\uparrow$ & gFID$\downarrow$ & gIS$\uparrow$ & Time & (\%) \\
    \hline
    VQGAN~\cite{llamagen} &{576}  & 0.95 & 197.3 & 3.71 & 228.3 & 4.3 & 23.1 \\   
   \hline
    VQGAN (CLIP) &\multirow{4}{*}{576} &  1.47 & 182.0  & 3.45 & 221.2 &4.0 & 59.5 \\ 
    VQGAN (SigLIP) & &  1.26 & 190.8  & 3.50 & 246.1 &  4.0  & 60.3\\  
    VQGAN (SigLIP2) & &  0.96 & 198.4  & 3.39 & 267.8 & 4.0 & 55.5 \\ 
    VQGAN (DINOv2) & &  0.99 & 206.3  & 3.34 & 268.6 & 4.0& 56.4 \\ 
    \hline
    VFMTok (CLIP) & \multirow{4}{*}{256} & 0.99 & 200.1 & 3.40 & 274.7 & 1.0 & 63.9 \\ 
    VFMTok (SigLIP) &  & 0.98 & 214.5 & {2.92} & {276.0} & 1.0 & 78.6 \\ 
    VFMTok (SigLIP2) &  & 0.94 & \textbf{218.7} & \textbf{3.01} & \textbf{280.8} & 1.0 & \textbf{78.5} \\ 
    VFMTok (DINOv2) & ~ & \textbf{0.89} & {215.4} & {3.08} & {274.2} & 1.0 & 69.4 \\ 
    \Xhline{0.8pt}
	\end{tabular}}
    \vspace{-.4cm}
\end{table}

%% file: sections/2_related_work.tex
\section{Related Work}
\label{related_work}

\noindent\textbf{Vision Foundation Models.} Vision Foundation Models (VFMs)~\cite{resnet,dino,dinov2, byol, he2019moco, chen2020mocov2,clip,align,siglip,siglip2} aim to learn general, transferable visual representations from large-scale, diverse data. The training of these versatile models has shifted from early supervised approaches to more scalable self-supervised learning~\cite{dino, byol, he2019moco, chen2020mocov2, bert, mae, dinov2, dinov2reg}, which leverages inherent data structures. More recently, language-supervised pre-training~\cite{siglip, align,siglip2} on vast image-text pairs has enabled VFMs to learn rich, semantically grounded representations. Pre-trained VFMs serve as powerful backbones for a wide array of downstream tasks. In this work, we utilize pre-trained VFMs directly as image tokenizers for AR image generation, surpassing other methods~\cite{vqgan-lc, zhu2024stabilize} with superior performance. Furthermore, using VFMs as tokenizers enables the removal of classifier-free guidance.

\noindent\textbf{Discrete Image Tokenization.} Pixel-space images are highly redundant. Autoencoder-based tokenizers~\cite{magvit, magvit2, var, llamagen} create compact latent tokens to reduce redundancy. VQVAEs~\cite{vqvae, vqvae2, kingma2013auto} and their derivatives evolved using adversarial losses~\cite{vqgan}, Transformers~\cite{vit-vqgan}, multistage quantization~\cite{residual-vq, movq}, lookup-free methods~\cite{magvit2, fsq}, and codebook initialization from pre-trained features~\cite{vqgan-lc, zhu2024stabilize}). These 2D tokenizers map features to a static 2D grid, which limits redundancy exploration. Recent 1D tokenizers~\cite{flextok, titok, oneD, semanticist} offer superior compression, reconstruction, and redundancy removal, but often require complex and lengthy training. For example, TiTok~\cite{titok} requires a two-stage process (warming up and fine-tuning) for 200 epochs. Our VFMTok adopts a novel region-adaptive tokenization framework to reduce redundancy. With a simpler training strategy for only 50 epochs, VFMTok exhibits discriminative semantics and excellent generation results. 

\noindent\textbf{Continuous-valued Image Tokenization.} Traditional Autoencoders and VAEs~\cite{ae, vincent2010stacked, diederik2019introduction, kingma2013auto, beta-vae} compress images into low-dimensional latent spaces. Subsequent advances have explored structure preservation (EQVAE~\cite{eq-vae}), scaling behaviors (ViTok~\cite{hansen2025learnings}), and integrating generative priors (REPA-E~\cite{repae}). To further improve latent learning, recent studies incorporate semantic representations from vision foundation models (VFMs). For instance, VA-VAE~\cite{va-vae} employs distillation to align its latent space with VFMs, while UniLIP~\cite{unilip} initializes its encoder with pre-trained CLIP~\cite{clip} and further optimizes it through distillation from a frozen CLIP as teacher. However, these approaches utilize VFMs merely as distillation targets or necessitate extensive fine-tuning, assuming frozen encoders lack visual details.
While recent works such as RAE~\cite{rae}, FAE~\cite{fae}, and VFMVAE~\cite{vfm-vae} explore building tokenizers directly on top of frozen VFMs, they suffer from notable limitations. By reconstructing images from uncompressed high-dimensional features, RAE~\cite{rae} yields sub-optimal reconstruction quality compared to standard continuous-valued counterparts. Besides, its limited versatility necessitates an intricate quantization strategy or architectural redesigns to accommodate autoregressive (AR) generation. Meanwhile, to achieve competitive performance, FAE~\cite{fae} relies on a cumbersome multi-stage alignment and fine-tuning pipeline, whereas VFMVAE~\cite{vfm-vae} resorts to excessively complex decoder architectures.

In contrast, our proposed VFMTok builds directly on top of a frozen VFM by utilizing a highly simplified architecture. Coupled with a joint objective that simultaneously reconstructs both raw images and semantic features, VFMTok efficiently compresses representations into low-dimensional spaces without the need for complicated multi-stage training. Consequently, it emerges as a versatile, generalist visual tokenizer seamlessly compatible with both discrete and continuous-valued image reconstruction and generation, delivering strong performance across both tasks. Beyond merely utilizing VFMs to facilitate image generation, we also systematically investigate the underlying mechanisms of why VFMs benefit the generative models.

\noindent\textbf{Denoising Image Generation.}  Historically, the field of visual synthesis was predominantly driven by continuous-valued generative frameworks. Generative Adversarial Networks (GANs)\cite{goodfellow2020gan,mirza2014cgan,radford2015dcgan,salimans2016improvedgans,isola2017Pix2Pix,zhu2017CycleGAN, karras2019stylegan} significantly enhanced visual realism. Despite these successes, the broader application of GANs remains fundamentally constrained by inherent training instabilities and susceptibility to mode collapse. To address these limitations, iterative denoising frameworks~\cite{ddpm, nichol2021iddpm, song2020denoising,dhariwal2021diffusion, liu2022flow, albergo2022building, flow-matching} have recently emerged as the prevailing paradigm. By formulating generation as a progressive denoising process, these models achieve superior sample quality and diversity. Furthermore, large-scale implementations such as Imagen~\cite{saharia2022Imagen} and Stable Diffusion~\cite{sd2.0, sd3.0} have propelled text-to-image synthesis to unprecedented levels of performance.

\noindent\textbf{Autoregressive Image Generation.} GPT-style Transformers~\cite{gpt, chen2020generative, pixelgpt, llamagen, residual-vq, var} have spurred interest in autoregressive (AR) image generation, which predicts visual token sequences. While early AR models operated in pixel space~\cite{chen2020generative, pixelgpt}, current methods~\cite{residual-vq, vit-vqgan, var, llamagen} generate discrete latent tokens via next-token prediction, then decode them to pixels using a tokenizer's decoder~\cite{vqvae, vqvae2, vit-vqgan, vqgan}. To improve the generation quality, recent works~\cite{var, mar, show-o} add bidirectional attention (\eg, VAR's next-scale prediction~\cite{var}, MAR's BERT-style framework~\cite{mar}, Show-o's hybrid attention~\cite{show-o}). These innovations, however, complicate designing universal, multi-modal Transformers adhering to next-token prediction. Instead, our {\ours} enables standard AR transformers to generate contextual token sequences for subsequent decoding, eliminating complex structural modifications. 

%% file: sections/3_approach.tex
\section{Method}

In this section, we first review the background of image tokenizers and present pilot studies on leveraging vision foundation models (VFMs) for tokenization. Building on these insights, we introduce \ours, a novel tokenizer built upon a frozen VFM. By incorporating region-adaptive strategy, \ours significantly enhances both the efficiency and representation quality of the tokenization process.

\subsection{Preliminary}
\label{method:preliminary}

\noindent\textbf{Image Tokenization.} \label{vq-procedure}
To perform high-fidelity visual generation efficiently, modern generative frameworks---such as autoregressive (AR) generative models~\cite{vit-vqgan,yu2022scaling,llamagen,var} or iterative denoising models ~\cite{sd2.0, flow-matching, liu2022flow}---necessitate an image tokenizer to convert raw pixels into a compact latent space. This mapping is fundamentally built upon autoencoder architectures, which can be categorized into \textit{discrete quantized tokenizers} (e.g., VQGANs~\cite{vqvae, vqgan, vit-vqgan}) for AR generation, and \textit{continuous-valued tokenizers} (e.g., standard VAEs~\cite{kingma2013auto, beta-vae, undertanding_beta-vae}) for denoising models.

Typically, an autoencoder-based tokenizer consists of an encoder $\mathcal{E}(\cdot)$, a latent bottleneck module $\mathcal{B}(\cdot)$, and a decoder $\mathcal{D}(\cdot)$. Given an input image $\text{I} \in \mathbb{R}^{H \times W \times 3}$, the encoder $\mathcal{E}(\cdot)$ first converts the image into spatial patch embeddings $Z_{2D} \in \mathbb{R}^{\frac{H}{f} \times \frac{W}{f} \times \text{D}}$, where $f$ is the spatial down-sampling factor and $D$ depicts the latent feature dimension. 

Subsequently, $Z_{2D}$ is processed by the bottleneck module $\mathcal{B}(\cdot)$ depending on the tokenizer type. For \textit{continuous-valued tokenizers}, $\mathcal{B}(\cdot)$ typically imposes a regularization penalty (e.g., KL-divergence~\cite{kingma2013auto} towards a standard Gaussian prior) or applies a simple scaling factor~\cite{ae} to yield a continuous latent representation $\hat{Z}$. Conversely, for \textit{discrete quantized tokenizers}, $\mathcal{B}(\cdot)$ acts as a quantizer $\mathcal{VQ}(\cdot)$ that maintains a learnable codebook $\mathbb{C} \in \mathbb{R}^{N \times D}$ with $N$ vectors. Each feature vector ${z_i} \in \mathbb{R}^{D}$ from $Z_{2D}$ is mapped to its nearest vector ${c_i} \in \mathbb{R}^{D}$ in the codebook $\mathbb{C}$. The general encoding and specific quantization processes can be formulated as:
\begin{equation}
\begin{split}
{Z_{2D}} &= \mathcal{E}(\text{I}) \,, \\
\mathcal{VQ}({z_i}) = {c_i}, \quad \text{where} \quad {c_i} &= \mathop{\arg\min}\limits_{{c_j} \in \mathbb{C}} \Vert {z_i} - {c_j} \Vert_2 \,.
\end{split}\label{equ:vq}
\end{equation}

Once the latent representations $\hat{Z}$ (either continuous features or de-quantized discrete codes) are acquired, they are converted back to the image pixel space by the decoder $\mathcal{D}(\cdot)$, as depicted in \cref{equ:dec}:
\begin{equation}
\label{equ:dec}
\mathbf{\hat{I}} = \mathcal{D}(\hat{Z}) \,.
\end{equation}

To optimize such autoencoders, the overall training objective generally consists of a shared reconstruction loss and a specific latent regularization loss. For high-fidelity image reconstruction, both continuous and discrete tokenizers commonly adopt the following objective:
\begin{equation}
    \mathcal{L}_{AE} = \mathcal{L}_2(\text{I}, \hat{\text{I}}) + \mathcal{L}_{P}(\text{I}, \hat{\text{I}}) + \lambda_{G} \cdot \mathcal{L}_{G}(\hat{\text{I}})\,,
\end{equation}
where $\mathcal{L}_2$ is a pixel-wise reconstruction loss, $\mathcal{L}_{P}$ is the perceptual loss derived from LPIPS~\cite{percetualloss}, and $\mathcal{L}_{G}$ is the adversarial loss from PatchGAN~\cite{patchgan} or DINOGAN~\cite{dino} governed by the weight $\lambda_{G}$. Additionally, specific latent regularizations are applied: \textit{continuous-valued tokenizers} rely on KL-divergence, whereas \textit{quantized tokenizers} optimize the codebook via $\mathbf{\mathcal{L}_{vq}}=\sum{\|\mathbf{sg}(z_i) -{c_i}\|^{2}_{2} + {\beta} \cdot \| \mathbf{sg}({c_i}) - {z_i}\|^{2}_{2}} $, where $\mathbf{sg}(\cdot)$ represents the stop-gradient operation~\cite{straightsthrough,vqvae}, and $\beta$ weights the commitment loss to align extracted features with the codebook vectors.

\subsection{Pilot Study: Pre-trained Vision Foundation Models as Tokenizers for AR Generation}
\label{method:pilot}
To investigate the feasibility of using pre-trained VFMs as image tokenizers, we conduct a pilot study. Specifically, we extract 2D grid features from $336\times336$ images using frozen VFMs (e.g., DINOv2~\cite{dinov2reg}, CLIP~\cite{clip}, and SigLIP2~\cite{siglip2}). After quantization, these features are fed into a VQGAN~\cite{llamagen} decoder for image reconstruction. We train the tokenizer for 50 epochs, and subsequently train a LLaMA-based AR model~\cite{llamagen} on the resulting discrete tokens for 100 epochs to evaluate downstream image synthesis.

As illustrated in \cref{tbl:prelimiary_exp}, directly utilizing frozen VFM features yields highly competitive reconstruction and generation performance compared to vanilla VQGANs. Most notably, these VFM-based tokenizers consistently demonstrate superior semantic representation capabilities, as evidenced by linear probing experiment. For instance, VQGAN (SigLIP2) matches the reconstruction quality of the vanilla baseline while delivering richer semantics and better generation results. However, performance varies significantly depending on the choice of VFM. While VQGAN(DINOv2) and VQGAN(SigLIP2) outperform vanilla VQGAN, VQGAN(CLIP) trails vanilla VQGAN. One contributing factor is that different learning objectives used to train VFMs influence their ability to extract detailed and semantic features from images, thereby affecting downstream image reconstruction and generation quality. As evidence, both DINOv2~\cite{dinov2reg} and SigLIP2~\cite{siglip2} employed a masked prediction objective to optimize their VFMs, whereas CLIP~\cite{clip} did not.

\subsection{VFMTok}

Building upon the semantically rich features provided by vision foundation models (VFMs)---which are typically structured as rigid 2D grids---\ours introduces a region-adaptive tokenization strategy. It dynamically identifies semantically coherent, irregular local regions to produce adaptive tokens. To support diverse generative paradigms, \ours can function as either a \textit{discrete tokenizer} (through vector quantization) or a \textit{continuous AutoEncoder} (with direct dimensional reduction). In the following sections, we describe the architecture of \ours, including its region-adaptive token generation module, the latent bottleneck, and the decoder for both image and VFM feature reconstruction. Besides, we outline the training objectives, which couple a pixel-level reconstruction loss for high-fidelity image reconstruction with a feature-level reconstruction loss to ensure semantic alignment.

\begin{figure*}[thbp]
  \centering
  \includegraphics[scale=0.5]{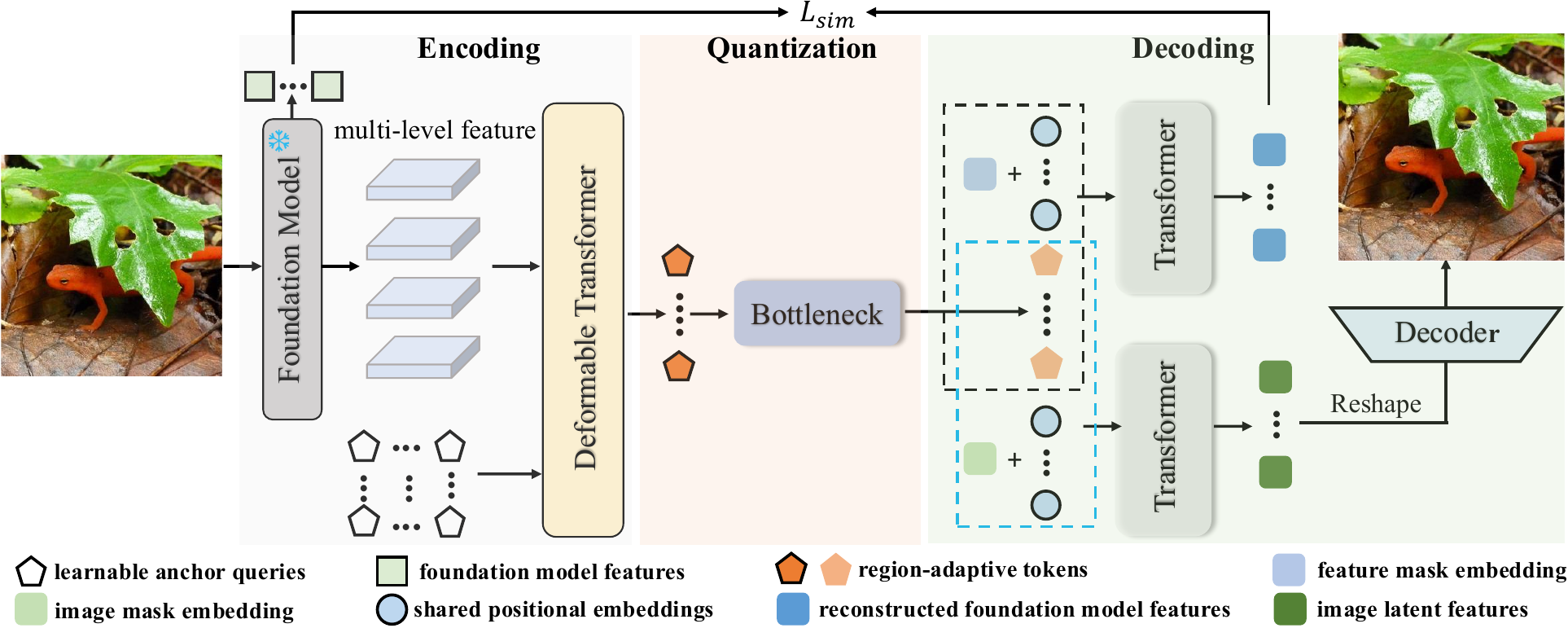}
  \vspace{-.3cm}
  \caption{The framework of \ours/VFMAE. \ours/VFMAE utilizes a frozen VFM to extract multi-level image features. A deformable Transformer then processes these features with learnable grid queries to generate region-adaptive tokens. After passing through the Bottleneck (either quantizer in discrete VQGAN or bottleneck layer in continuous-valued AE), these tokens are fed into Transformers to perform dual reconstruction: 1) VFM features, targeting similarity with the VFM's last-layer outputs, and 2) image latent features, which are reshaped to a 2D grid and decoded into pixels.}
  \label{fig:framework}
  \vspace{-0.55cm}
\end{figure*}

\noindent\textbf{Region-adaptive Token Generation.}
Following our pilot study, we utilize a frozen pre-trained vision foundation model (VFM) to encode an input image $I$ into latent embeddings  $\mathcal{F}$. Since VFM features encode crucial fine-grained details in shallower layers and high-level semantics in deeper layers~\cite{opmp, fpn, retinanet}---both of which are essential to image reconstruction---we extract multi-level features $\mathcal{F_\text{m}}$ from the VFM. These multi-level features are then projected into a uniform embedding dimension using a two-layer MLP. Next, as illustrated in Fig.~\ref{fig:framework}, we apply a region-adaptive sampling mechanism over $\mathcal{F_\text{m}}$ using deformable cross-attention~\cite{deformable_conv, deform_detr}. Specifically, a set of learnable anchor queries (initialized as a 2D grid) is iteratively refined across $d_1$ deformable attention layers. In each layer, an anchor query uses a linear projection to predict sampling offsets for each VFM feature level, allowing it to sample from irregular, data-dependent spatial locations. These sampled features are then weighted by attention scores and aggregated to update the query. Through this process, the anchor queries capture semantically coherent, region-specific information, resulting in a set of continuous region-adaptive features $Z_r$. Compared to standard fixed 2D grids, this adaptive aggregation drastically reduces spatial redundancy, enabling high-fidelity reconstruction and generation with significantly fewer tokens (\textit{e.g.}, 576 vs. 256 tokens, as shown in Tab.~\ref{tbl:prelimiary_exp}).

\noindent\textbf{Latent Bottleneck: Discrete vs. Continuous Spaces.} \label{sec:quantization}
Depending on the downstream generative paradigms, the region-adaptive features $Z_r$ are processed into the final latent tokens $\hat{Z}_r$ using one of two distinct bottlenecks:
\begin{itemize}
    \item \textbf{Discrete Tokenizer (VQGAN):} To produce discrete codes $\Tilde{Z}_r$, the feature $Z_r$ is first compressed into a low-dimensional feature space, then a quantizer $Q_c(\cdot)$ discretizes it using a learnable codebook. Following the practices in~\cite{vit-vqgan, llamagen}, we apply $\ell_2$-normalization to the codebook vectors and utilize a low-dimensional embedding space coupled with a large codebook size to maximize codebook utilization and reconstruction quality. In this case, $\hat{Z}_r$ are discrete tokens.
    
    \item \textbf{Continuous AutoEncoder:} To function as a continuous-valued tokenizer (e.g., for denoising generative models), $Z_r$ is directly projected into low-dimensional continuous features with a linear projection to form the latent tokens, $\hat{Z}_r$.
\end{itemize}

\noindent\textbf{Decoder for Image and VFM Feature Reconstruction.}
Regardless of the bottleneck choice, the resulting latents $\hat{Z}_r$ must be spatially aligned back into a regular 2D grid for decoding. To achieve this, we introduce a mask token sequence $M_\text{I} \in \mathbb{R}^{H_m \times W_m \times C}$, initialized by replicating a single learnable token and adding spatial position embeddings $E$. 

The latent features $\hat{Z}_r$ are first up-projected with a one-layer MLP to increase their dimensionality, and then concatenated with $M_\text{I}$ and processed by a lightweight Transformer $\mathcal{E}_\text{ViT}(\cdot)$  of depth $d_2$, which propagates information from $\hat{Z}_r$ to the regular masked grid. Following DINOv2~\cite{dinov2reg}, we append a $\verb+[CLS]+$ token and register tokens $\text{Tok}_\text{reg}$ to improve global context modeling, though these are not used for reconstruction. The output corresponding to the mask tokens, denoted as $\mathcal{F}_I$, is reshaped into a spatial grid and passed through a decoder $\mathcal{D}(\cdot)$ to reconstruct the original image.

To explicitly preserve the semantic integrity of the latents, we concurrently reconstruct the high-level features from the VFM's final layer. This process mirrors the image reconstruction: a distinct set of mask tokens ${M}_f \in \mathbb{R}^{H_m \times W_m \times C}$ (sharing the same position embeddings $E$) is concatenated with $\hat{Z}_r$. The combined sequence is processed by another Transformer $\mathcal{\hat{E}}_\text{ViT}(\cdot)$ of depth $d_3$ to predict the VFM feature map $\mathcal{F}_\text{P}$. Note that VFM feature reconstruction is only enforced during tokenizer training.

\noindent\textbf{Training Objective.}
For tokenizer optimization, we adopt the standard VQGAN~\cite{vqgan, llamagen} framework with a crucial modification: we replace the conventional PatchGAN discriminator~\cite{patchgan} with a pre-trained DINOv1-S~\cite{dino} model. This provides adversarial guidance in a semantically rich feature space, consistently yielding superior reconstruction quality. Furthermore, to guide the feature reconstruction branch, we apply a cosine similarity loss between the predicted features $\mathcal{F}_\text{P}$ and the frozen VFM targets. The overall objective is formulated as $\mathcal{L} = \alpha \cdot \mathcal{L}_\text{AE} + \lambda \cdot \mathcal{L}_{\text{sim}}$, where $\mathcal{L}_\text{AE}$ is the image reconstruction loss and $\mathcal{L}_{\text{sim}}$ is the feature similarity loss. We empirically set both $\alpha$ and $\lambda$ to 1.

\subsection {Autoregressive Image Generation}
Once \ours is trained in the discrete setting, the optimized region-adaptive tokens $\Tilde{Z}_r$ can be seamlessly integrated into an autoregressive (AR) Transformer. Conditioned on a class or text embedding $c$, tokens are generated sequentially via next-token prediction. Notably, to better capture spatial locality during AR modeling, the region-adaptive tokens are augmented with 2D Rotary Position Embeddings (RoPE)~\cite{rope} before being rendered into images using $\mathcal{E}_{\text{ViT}}(\cdot)$ and $\mathcal{D}(\cdot)$.

\subsection{Denoising Image Generation}
When configured as a continuous autoencoder, \ours provides a highly structured latent space where the region-adaptive tokens $\tilde{Z}_r$ can be natively synthesized by a denoising generative model. Conditioned on a class or text embedding $c$, the generation process starts by sampling pure Gaussian noise in the continuous latent space. A denoising network~\cite{ddpm, flow-matching} then iteratively removes the noise over a series of timesteps to produce the clean continuous tokens $\tilde{Z}_r$. Finally, mirroring the reconstruction phase, these generated tokens are concatenated with the mask tokens $M_I$, processed by the shared Transformer $\mathcal{E}_{\text{ViT}}(\cdot)$ to recover the spatial grid, and ultimately rendered into high‑fidelity pixels by $\mathcal{D}(\cdot)$.

%% file: sections/4_experiments.tex
\section{Experiment}
\subsection{Image Tokenizer Setup} \label{sec:setup}

\noindent\textbf{Discrete Image Tokenizer.}
In our main experiments, we initialize the encoder of \ours with a frozen pre-trained DINOv2-L~\cite{dinov2reg}. To capture multi-level representations from its 24 Transformer layers, we extract features from the 6th, 12th, 18th, and 24th layers. Each image is encoded into exactly 256 region-adaptive tokens. Following established practices~\cite{llamagen, titok}, we configure the quantizer with a codebook dimension of 12 and a codebook size of 16,384 to ensure optimal reconstruction quality and efficient codebook utilization. Besides, the depths of the Transformer modules ($d_1$, $d_2$, and $d_3$) are uniformly set to 6 (following~\cite{deform_detr}). Additionally, the Transformers, $\mathcal{E}_\text{ViT}(\cdot)$ and $\hat{\mathcal{E}}_\text{ViT}(\cdot)$, share parameters across both image and VFM feature reconstruction, effectively reducing the tokenizer's parameter footprint. Additionally, the number of register tokens $\text{Tok}_\text{reg}$ is set to 4, as consistent with that in DINOv2~\cite{dinov2reg}.

The tokenizer is trained on the ImageNet~\cite{imagenet} training set. Since standard pre-trained VFMs~\cite{dinov2reg, clip, align, siglip, siglip2} typically operate at a resolution of $336\times336$, we train \ours at this native resolution. Otherwise, our training hyperparameters strictly follow LlamaGen~\cite{llamagen}. For evaluation on the ImageNet validation set, we ensure a fair comparison with vanilla tokenizers~\cite{vqgan, vqgan-lc, llamagen} by adopting LlamaGen's evaluation protocol: images are reconstructed at $336\times336$ and subsequently resized to $256\times256$ before calculating the evaluation metrics.

\noindent\textbf{Continue-valued AutoEncoder.}
Compared to its discrete counterpart, the continuous-valued \ours introduces four key modifications. First, the channel dimension of the compressed latent space is set to 32, matching that of VA-VAE~\cite{va-vae}. Second, while the Transformer depths $d_1$ and $d_2$ remain at 6, $d_3$ is reduced to 1. This shallow design ensures that the semantic reconstruction loss directly supervises the low-dimensional latent features, preserving richer structural semantics. Consequently, it allows downstream generative models to learn the latent spatial distribution more efficiently. Third, to enrich the multi-level representations with global context, the $\verb+[CLS]+$ token from each specific layer is added element-wise to its relevant spatial feature map. While the register tokens are removed from the learnable queries sequence. Last, to guarantee fair comparisons with baseline denoising models~\cite{flow-matching, liu2022flow}, our continuous-valued \ours is trained and evaluated natively at $256\times256$ solution on ImageNet. All other configurations remain identical to the discrete \ours. Considering the continuous VFMTok is essentially an AutoEncoder, we also dub it as VFMAE.

\subsection{Generative Model Setup}
\noindent\textbf{Class-conditional Autoregressive Image Generation.} 
Following the protocol in LlamaGen~\cite{llamagen}, our AR models generate images at a resolution of $336\times336$, which are subsequently resized to $256\times256$ for evaluation. We conduct experiments across a range of model scales, from 111M to 3.1B parameters, to assess the scalability of our proposed image tokenizer. To balance computational costs, training durations are scaled according to model size: models with fewer than 1B parameters are trained for 300 epochs, while larger models are trained for 200 epochs. Aside from resolution and training duration, all hyperparameters strictly follow LlamaGen~\cite{llamagen}. To demonstrate generalizability, we also integrate \ours into an advanced RAR~\cite{rar} framework, adopting its default training configurations. During inference, we evaluate our AR models under both classifier-free guidance (w/ CFG) and without classifier-free (w/o CFG) guidance settings.

\noindent\textbf{Class-conditional Denoising Image Generation.}
We incorporate the continuous-valued VFMTok into the denoising generative model, $\text{DiT}^\text{DH}\text{-XL}$~\cite{rae}, whose architectural configuration strictly follows that established in RAE~\cite{rae}. Additionally, we also utilize the noise schedule strategy proposed in ~\cite{esser2024scaling}. Concretely, for a schedule $t_n \in [0, 1]$ and input dimension $n, m$, the shifted timestep is defined as $t_m =\frac{\alpha{t_n}}{1 + (\alpha - 1){t_n}}$, where $\alpha = \sqrt{m/n}$ is dimension-dependent scaling factor. Here we follow~\cite{esser2024scaling} in using $n=4096$ as the base dimension and set $m$ equals 65,536 by default. In terms of the training protocol, the model is trained on the ImageNet training set and estimated on the validation set at a standard resolution of $256\times256$. The training process spans a total of 800 epochs to ensure consistency with mainstream denoising models~\cite{ddpm, flow-matching,sit,fae,vfm-vae}. During inference, following RAE~\cite{rae}, AutoGuidance~\cite{autoguidance} is also adopted to ensure a fair comparison. Additionally, we report the generative performance evaluated at 80 and 800 epochs separately.

\subsection{Evaluation Metrics} 
To comprehensively evaluate image reconstruction, we adopt reconstruction Fréchet Inception Distance (rFID~\cite{fid}) and reconstruction Inception Score (rIS~\cite{is}) across all tokenizers. For discrete tokenizers, we also measure codebook utilization. For continuous tokenizers, we supplement the evaluation with PSNR and SSIM~\cite{ssim} to estimate reconstruction fidelity. For generative performance, each model synthesizes 50,000 images spanning all 1,000 ImageNet~\cite{imagenet} classes, which are subsequently evaluated using generation Fréchet Inception Distance(gFID~\cite{fid}), generation Inception Score (gIS~\cite{is}), and Precision as well as Recall~\cite{pre_recall}. Because the fundamental definitions of rFID/gFID and rIS/gIS are identical, the unified definitions for FID, IS, and the remaining metrics are introduced below.

\begin{itemize}
    \item \textbf{Fréchet Inception Distance (FID)} measures the similarity between the distributions of real and generated images in the feature space of InceptionV3~\cite{inception-net}. A lower FID score indicates higher visual quality and better fidelity to the training data distribution.
    
    \item \textbf{Inception Score (IS)} assesses both the quality (clarity) and diversity of generated images. A higher IS suggests that the model generates more distinct and recognizable objects.
    
    \item \textbf{Peak Signal-to-Noise Ratio (PSNR)} computes the pixel-level absolute error by measuring the ratio between the maximum possible signal power and the distorting noise. A higher PSNR value indicates lower reconstruction error and better pixel-level fidelity to the original image.

    \item \textbf{Structural Similarity Index Measure (SSIM)} evaluated the perceived visual quality by comparing the luminance, contrast, and structural dependencies between the original and reconstructed images. A higher SSIM score (closer to 1) indicates better preservation of structural information and greater visual similarity to the original image.
    
    \item \textbf{Precision and Recall} are used to estimate class-conditional generation. Precision measures the fidelity of generated samples (what fraction are realistic), while Recall measures diversity (what fraction of the real data distribution is covered). Higher values for both are desirable.
        
\end{itemize}

\input{tables/discrete_tok_eval}

\subsection{Main Results}

\noindent\textbf{Discrete Image Reconstruction.}We build discrete VFMTok using different frozen pre-trained VFM (DINOv2-L~\cite{dinov2reg}, SigLIP-L~\cite{siglip}, and SigLIP-L~\cite{siglip2}) for image reconstruction. Then we compare \ours against representative 2D image tokenizers, including VQGAN~\cite{vqgan}, MaskGiT~\cite{maskgit}, ViT-VQGAN~\cite{vit-vqgan}, as well as a 1D tokenizer, TiTok~\cite{titok}. As shown in~\cref{tbl:tok_comparison2}, our tokenizer represents an image with just 256 tokens, considerably fewer than some counterparts. For instance, the VQGAN variant LlamaGen~\cite{llamagen} uses 576 tokens, while VQGAN~\cite{vqgan} and ViT-VQGAN~\cite{vit-vqgan} even utilize up to 1024 tokens. Despite this efficiency, {\ours} achieves a strong rFID of \textbf{0.89}, and further demonstrates 100\% utilization of the codebook. Besides, the rIS score of \textbf{215.4} achieved by \ours significantly outperforms other methods, \eg, TiTok~\cite{titok} and the VQGAN series. The smaller rFID and higher rIS confirm \ours is more effective at preserving semantic consistency during reconstruction.

\noindent\textbf{Continuous-valued Image Reconstruction.}
\cref{tbl:vfmae_img_recon} compares VFMAE with existing continuous-valued image tokenizers on the ImageNet~\cite{imagenet} validation set. A key observation is the efficiency of our latent space. While RAE employs a large channel dimension of 768 to achieve an rIS of 226.8, our VMFAE(DINOv2) and VFMAE(SigLIP2) achieve comparable rIS scores (225.6 and 225.4, respectively) using a substantially more compact dimension of 32, which facilitates the fitting of the latent distribution for denoising generative models, as denoising models are known to struggle in high-channel latent spaces due to unstable optimization and less effective noise scheduling~\cite{rae, aligntok}. Furthermore, VFMTok surpasses RAE by a large margin in terms of rFID (0.29 vs. 0.57). When compared to the approaches sharing a similar philosophy of using VFM to build an image tokenizer, such as VFMVAE~\cite{vfm-vae} and AlignTok~\cite{aligntok}, our VMFAE (DINOv2) maintains competitive rFID (0.29) and PSNR (25.5), which validates the effectiveness of incorporating pre-trained vision foundation models for image tokenization.

\input{tables/continuous_tok_eval}
\input{tables/whole_discrete_gen}

\noindent\textbf{Autoregressive Class-conditional Image Generation.} We evaluate VFMTok built upon different VFM (DINOv2-L~\cite{dinov2reg}, SigLIP-L~\cite{siglip2}, SigLIP2-L~\cite{siglip2}) on vanilla autoregressive models -- LlamaGen~\cite{llamagen}, and advanced generative model -- RAR~\cite{rar} with different scales by conducting  $256\times256$ class-conditional image generation task on ImageNet~\cite{imagenet}, where comparing them with the mainstream generation models, including diffusion models (Diff.)~\cite{maskDiT, peebles2023scalable, sit, fasterdit}, BERT-style masked-prediction models (Mask.)~\cite{maskgit}, and AR generation models (AR)~\cite{titok, vqgan, vit-vqgan, residual-vq, vqvae2, llamagen}. 
As shown in \cref{tbl:distict-gen}, our models exhibit competitive performance across all metrics compared to mainstream image generation models. Notably,  \ours beats BERT-style models~\cite{maskgit} in terms of gFID without the requirement of complicated sampling tuning. With comparable or even fewer parameters, our method surpasses most AR generative models~\cite{titok, vqgan, vit-vqgan, residual-vq, vqvae2} in both gFID and gIS metrics. Under the same training setting, \ours surpasses LlamaGen~\cite{llamagen} by significant gFID gains and notable gIS improvements. Specifically, \ours-B outperforms LlamaGen-B~\cite{llamagen} with gains of \textbf{2.56} in gFID and \textbf{69.7} in gIS. Besides, our \ours-L model achieves a gFID of \textbf{2.75} at 300 epochs, also obtaining a gain of \textbf{22.7} in gIS. Notably, when compared with LlamaGen-3B with 3B parameters, our {\ours}-XXL (2.2B) achieves even better generation performance with less than half the number of parameters and fewer training iterations. Furthermore, when VFMTok is incorporated into RAR~\cite{rar}, it achieves a generative performance with gFID of~\textbf{1.36}, which is the state-of-the-art generation performance at present. Additionally, discrete class-conditional image generation results are visualized in the Appendix.

Furthermore, we conducted experiments by \textbf{removing classifier-free guidance (CFG)}. Remarkably, the generation results without CFG show that most evaluation metrics—such as sFID, Precision, and Recall—remain comparable to those obtained with CFG. While gIS experiences a slight decline, gFID improves compared to its CFG-enabled counterpart. Similar trends are also observed when VFMTok's encoder is replaced with other frozen pre-trained vision foundation models like SigLIP2~\cite{siglip2}.
These results demonstrate that our method supports high-fidelity autoregressive image generation even without CFG, which significantly accelerates inference. In contrast, baseline methods suffer substantial performance degradation without CFG---for example, LlamaGen-3B~\cite{llamagen} model witnesses gFID worsen to \textbf{9.38}, whereas our 1.4B model VFMTok-XXL achieves a gFID of \textbf{1.95} without CFG.

\input{tables/continuous_gen_eval} 
\noindent\textbf{Denoising Class-conditional Image Generation.}
\cref{tbl:continue_gen2} highlights the generation capabilities of continuous-valued tokenizers built upon a frozen pre-trained VFM (DINOv2-L~\cite{dinov2reg} and SigLIP2-L~\cite{siglip2}) on the ImageNet~\cite{imagenet} benchmark, explicitly illustrating the advantages of integrating vision foundation models (VFMs). By leveraging representations from either DINOv2~\cite{dinov2reg} or SigLIP2~\cite{siglip2}, our VFMAE consistently outperforms conventional VAEs~\cite{sd2.0}, while also matching or exceeding the performance of recent models (\textit{e.g.}, FAE~\cite{fae}, RAE~\cite{rae}, AlignTok~\cite{aligntok}, VFMVAE~\cite{vfm-vae}) that build tokenizers upon VFMs. As shown in \cref{tbl:continue_gen2}, when employing classifier-free guidance (w/ CFG), VFMAE (DINOv2) achieves an outstanding gFID of \textbf{1.25} after a standard 800-epoch training schedule. This not only surpasses existing diffusion models (\textit{e.g.}, MaskDiT~\cite{maskDiT}, DiT~\cite{peebles2023scalable}, SiT~\cite{sit}, and REPA~\cite{repa}) utilizing vanilla VAEs but also competes favorably with denoising models that adopt VFMs as their underlying tokenization backbones. Even in the settings without classifier-free guidance (w/o CFG), VFMAE demonstrates highly competitive performance against the strong counterparts, such as AlignTok~\cite{aligntok} and RAE~\cite{rae}.

Beyond generation quality, a critical advantage of VFMAE lies in its training efficiency. While denoising generative models~\cite{maskDiT,peebles2023scalable,sit,fasterdit,repa} trained on vanilla VAEs typically require extensive training duration of up to 1600 epochs, VFMAE delivers substantial improvements with only a standard schedule of 800 epochs. Compared to other VFM-driven tokenizers~\cite{rae, fae, aligntok, vfm-vae}, VFMAE maintains a distinct edge by consistently yielding comparable or superior generation fidelity. Most notably, VFMAE exhibits remarkable convergence speed: after merely 80 epochs of training, it already achieves an impressive gFID of 1.68. This strongly confirms that integrating VFMs significantly accelerates convergence while establishing a high-quality continuous-valued image synthesis.

Finally,  That by analyzing the reconstruction and generation results from \cref{tbl:vfmae_img_recon} and \cref{tbl:continue_gen2} reveals a critical insight. When building an image tokenizer upon a pre-trained VFM, introducing a semantic reconstruction objective to supervise the training effectively bypasses the requirement for high-dimensional latent spaces~\cite{rae} and complex multi-stage paradigms~\cite{fae}. Instead, through straightforward end-to-end optimization, this strategy produces a highly effective tokenizer that seamlessly facilitates superior generation quality in denoising generative models.

\subsection{Study of Components, Convergence and Efficiency}
\noindent\textbf{Core component analysis.}
To assess the contribution of each proposed component to image reconstruction and synthesis, we conduct a step-by-step component analysis using a baseline tokenizer built on vanilla VQGAN~\cite{llamagen}. We incrementally add the following components: (1) replace the VQGAN encoder with a frozen pre-trained foundation model (DINOv2-L~\cite{dinov2reg}); (2) introduce learnable queries and a deformable attention for region-adaptive tokenization, using only single-level features from the final layer; (3) incorporate multi-level features to enrich representations with both low-level detail and high-level semantics; and (4) add a feature reconstruction objective based on pre-trained VFM outputs.
After training each tokenizer, we integrate it with our AR generation model, \ours-L, for autoregressive image synthesis. Both the tokenizer and AR model are trained for 50 epochs. Additionally, we perform linear probing on the $\verb+[CLS]+$ token, following the MAE~\cite{mae} protocol.
\input{tables/core_comp_analysis}
As shown in \cref{tbl:ablation_queries}, replacing VQGAN’s encoder with a frozen pre-trained vision foundation model yields reconstruction and generation performance on par with a VQGAN trained specifically for visual reconstruction using 576 tokens. This substitution also significantly enhances the semantic quality of the tokenizer’s representations. To further improve token efficiency, we introduce region-adaptive tokenization using deformable attention to exploit the spatial redundancy inherent in regular 2D grid features. This reduces the number of visual tokens to 256. However, this performance gain comes at a cost: reconstruction and generation quality degrade slightly due to two factors: (1) fewer visual tokens limit representational capacity, and (2) the absence of explicit supervision hinders the effective optimization of the region-adaptive tokens. To address this, we incorporate multi-level feature extraction, which improves the reconstruction capability by leveraging both low- and high-level information. However, without additional guidance, the semantic consistency of the learned tokens may still degrade. Finally, we introduce a pre-trained feature reconstruction objective, which significantly boosts both image reconstruction and generation quality. This objective encourages alignment with semantic features from the frozen VFM and effectively balances the contributions of low- and high-level features to the contextual tokens—thereby preserving semantic fidelity.

With these three key components---(1) deformable attention for region-adaptive tokenization to reduce redundancy, (2) multi-level features for enhanced reconstruction, and (3) feature reconstruction loss for semantic alignment---VFMTok produces compact, semantically rich, and efficient tokens. Using only 256 tokens, VFMTok outperforms its VQGAN baseline with 576 tokens in reconstruction quality, generative performance, and semantic representation. Supplemental ablations are discussed in the Appendix.

\noindent\textbf{Convergence and efficiency analysis.}  
Beyond above analysis, we experiment VFMTok with a randomly initialized encoder instead of a pre-trained VFM with other components remaining unchanged. As shown in Tab.~\ref{tbl:ablation_queries} (last row), its reconstruction quality dropped to the level of VQGAN. Meanwhile, both its semantic representation capability and generation performance also decreased. This indicates a frozen VFM benefits tokenizer training as it provides a latent space advantageous for image reconstruction and generation. Besides, those semantic-rich, structured latent tokens accelerate AR model training convergence. As evidenced in Fig.~\ref{fig:pilot2}(b), VFMTok enables AR models to achieve a \textbf{3$\times$} speedup in convergence compared to VQGAN. Moreover, an AR model's generation time is quadratically proportional to the number of tokens. At the same resolution, VFMTok uses approximately half the tokens for image representation compared to counterparts like DINOv2-VQGAN and CLIP-VQGAN. Consequently, VFMTok achieves a \textbf{4$\times$} generation speedup over these counterparts depicted in Tab.~\ref{tbl:prelimiary_exp}. This acceleration can be further enhanced with CFG-free generation.

\noindent\textbf{Effect of Learnable Components.}
Our proposed \ours achieves exceptional image reconstruction and generation performance through three core designs: a frozen vision foundation model (\textit{e.g.}, DINOv2~\cite{dinov2reg}) as the encoder, multi-level feature interaction, and a pre-trained feature reconstruction objective. To implement these mechanisms, \ours incorporates several specialized learnable modules: distinct mask tokens ${M}_\text{I}$ and ${M}_{f}$ for image and feature reconstruction, deformable attention layers, and a set of (4) learnable register tokens $\text{Tok}_\text{reg}$ designed to mitigate potential artifacts in the latent space. Consequently, this architectural formulation naturally raises three critical design questions: 1) Can a single mask token ${M}_\text{shared}$ be shared between the image and feature reconstruction tasks? 2) Can the deformable attention (DA) be replaced by standard cross-attention (CA)? 3) Are the register tokens strictly necessary, or can they be entirely discarded from \ours?

To answer the questions raised above, we conducted 3 different experiments: 1) ${M}_\text{I}$ and ${M}_{f}$ share the same mask token ${M}_\text{share}$; 2) Replacing deformable attention with cross-attention in deformable transformer. To address the memory, computational demands, and fairness considerations of cross-attention, we first concatenate multi-level features from a VFM along the channel dimension, then apply a single MLP for dimensionality reduction before these features are fed to the cross-attention transformer for interaction with queries. Besides, each query interacts with VFM features within a $16\times16$ window to simulate region-adaptive behavior; 3) Removing the register tokens from \ours. Additionally, linear probing is carried out on the $\verb+[CLS]+$ token to estimate the semantic representation capability of \ours.

As shown in Tab.~\ref{tbl:part_annalysis}, unshared mask tokens seldom affect image reconstruction or generation quality but significantly degrade {\ours}'s overall semantic representation. This indicates that image reconstruction requires a certain amount of semantic information, but this semantic information is not as strong as that requested for VFM's feature reconstruction. Besides, Tab.~\ref{tbl:part_annalysis} also reveals that shared mask tokens can enhance the semantic representation of \ours, potentially benefiting downstream comprehension tasks. Hence, we use shared mask tokens by default in \ours. Introducing cross-attention for learnable queries and multi-level feature interaction shows an evident decline in image reconstruction, generation, and semantic level. Compared to deformable attention that focuses on local regions, using cross-attention may introduce redundant information, this redundancy weakens the overall semantic representation by complicating quantized visual token distribution, thereby hindering image generation. 
Additionally, we remove the learnable register tokens $\text{Tok}_\text{reg}$ from \ours. Interestingly, this removal only slightly impacts image reconstruction and generation, but causes a noticeable performance degradation in recognition accuracy (indicated by linear probing). This phenomenon suggests that the built-in register tokens of the DINOv2~\cite{dinov2reg} have already absorbed most spatial artifacts, leaving the dense features to benefit reconstruction. However, since the reconstruction objectives do not explicitly supervise the $\verb+[CLS]+$ token, the introduced register tokens remain indispensable for isolating residual noise and enhancing the robustness of context semantics extraction. Finally, given that register tokens have little impact on image reconstruction and generation, we directly removed them from  VFMAE.

\input{tables/feat_analysis_vfm_category}
\noindent\textbf{Effect of Single-level \textit{v.s.} Multi-level Features.} 
We also explore the distinct impacts of each single-level or multi-level feature on image reconstruction and generation. Specifically, we extract features from four equally spaced layers of DINOv2-L~\cite{dinov2reg}. Next, we perform discrete image reconstruction and autoregressive generation on each feature level individually. Subsequently, we utilize a cumulative strategy, starting with the lowest-level feature and progressively aggregating the remaining higher-level features.

We initialize and train each tokenizer for 50 epochs. Once optimized, it is integrated with an AR generation model, \ours-L, for a total of 100 training epochs. During evaluation, the performance of image reconstruction and generation is reported.

As presented in \cref{tbl:each-single-level}, the single-level analysis reveals a clear trade-off: features from the 2nd-level are highly beneficial for image reconstruction, whereas higher-level features yield superior generative performance but suffer from inferior reconstruction quality. Conversely, the cumulative experiments depicted in \cref{tbl:cumulative-features} illustrate that progressively introducing features level by level consistently improves both image reconstruction and generation. This strongly validates the necessity and effectiveness of incorporating multi-level features within the VFMTok design. Additionally, the observation from \cref{tbl:each-single-level} and \cref{tbl:cumulative-features} explicitly indicate that incorporating high-level semantic features is crucial for facilitating high-fidelity image generation.

\vspace{-0.3cm}
\subsection{What makes VFM a good visual tokenizer?}

To address this question, we systematically investigate how different VFM pre-training objectives—namely, masked image modeling in pixel or latent spaces (Pixel-MIM and Latent-MIM) and contrastive learning (C.L.)—affect the efficacy of VFMTok. By constructing tokenizers based on these distinct paradigms, we aim to provide deeper insights into the optimal feature space for image reconstruction and synthesis.

\noindent\textbf{Setup.} We replace the encoder of a standard VQGAN~\cite{llamagen} or AutoEncoder~\cite{ae} with various VFMs characterized by distinct pre-training objectives. Each resulting tokenizer is trained on the ImageNet~\cite{imagenet} training set for 50 epochs. Subsequently, these tokenizers are integrated into an autoregressive (AR) or denoising generative model, LlamaGen~\cite{llamagen} or $\text{DiT}^\text{DH}-\text{XL}$~\cite{rae}, and trained for an additional 100 epochs. Performance is evaluated on the ImageNet validation set, utilizing FID and IS to measure reconstruction and generation quality, respectively. Furthermore, we report the top-1 accuracy from linear probing to assess the semantic richness of the learned representations.

\noindent\textbf{Observations.} 
As depicted in \cref{tbl:mim_gen}, vision foundation models (VFMs) supervised solely by contrastive learning (C.L.), such as VQGAN(CLIP) and VQGAN(SigLIP), exhibit superior semantic representations (highest linear probing accuracy) but deliver modest reconstruction and generation fidelity. Conversely, models relying on Pixel-MIM (e.g., MAE~\cite{mae}) align most naturally with pixel-level reconstruction objectives, allowing VQGAN(MAE) to achieve optimal reconstruction quality. However, this objective fails to yield a latent space benefiting generative tasks, and its semantic expressiveness is inferior to other VFMs. Finally, latent-MIM and contrastive learning co-optimized VFM, VQGAN(SigLIP2), and VQGAN(DINOv2) achieve the best comprehensive performance across reconstruction, generation and semantic representation.

\input{tables/vfm_objectives}

\noindent\textbf{Discussion.} The training objectives of VFMs affect the image reconstruction and generation of VFMTok. \cref{tbl:vfms_objective_grouping} provides a holistic comparison of these objectives. CLIP~\cite{clip} and SigLIP~\cite{siglip} apply contrastive learning to image-text pairs, whereas DINO~\cite{dino} applies a contrastive learning-like clustering objective to images. MAE~\cite{mae} performs mask image modeling on image patches in pixel space, while iBOT~\cite{ibot} predicts DINO-like cluster assignments of image patches in latent space. DINOv2~\cite{dinov2reg} basically is the synergy of DINO~\cite{dino} and iBOT~\cite{ibot}, and SigLIP2~\cite{siglip2} is also essentially the integration of SigLIP~\cite{siglip} and iBOT~\cite{ibot} (termed TIPS loss ), omitting other regularizing losses.

Crucially, the mechanisms underlying iBOT~\cite{ibot}, DINOv2~\cite{dinov2reg}, and SigLIP2~\cite{siglip2} exhibit a profound conceptual alignment with the codebook learning paradigm of VQGAN. Although these VFMs utilize soft probabilistic distributions rather than discrete one-hot encodings and maintain high-dimensional codebook vectors (\textit{e.g.}, 256), the core quantization principle remains analogous. Following DINO~\cite{dino}, these models learn a massive vocabulary of prototypes (\textit{e.g.}, 8,192 or 65,536) and enforce consistent prototype assignments across different augmented views of the same image or patch. By further requiring the prediction of these soft assignments for masked patches, they effectively perform Latent-MIM. This formulation—originally introduced as self-distillation in DINO, extended to patch-level Latent-MIM in iBOT, and subsequently inherited by DINOv2~\cite{dinov2reg} and SigLIP2~\cite{siglip2}—serves as a continuous-space analog to discrete vector quantization. Additionally, the conceptual alignment between Latent-MIM and VQGAN makes these VFMs inherently suitable for visual tokenization, yielding robust image reconstruction and synthesis. Additionally, the global contrastive learning objective adopted in DINOv2 and SigLIP2 also promises better high-level semantics (better performance on understanding tasks).

To summarize, the conclusion can be drawn as: Contrastive learning (C.L.) objective is less helpful for reconstruction and generation, but is important for understanding abilities (\textit{e.g.}, top-1 accuracy on ImageNet~\cite{imagenet}). Masked image modeling (MIM) objectives primarily help reconstruction and generation. MIM in pixel space helps reconstruction more, but is less beneficial for generation than MIM in latent space. Best VFMs for visual tokenization are trained with both mask image modeling in latent space and contrastive objective, namely DINOv2~\cite{dinov2reg} and SigLIP2~\cite{dinov2reg}.

%% file: tables/discrete_tok_eval.tex
\begin{table}[thbp]
\vspace{-0.5cm}
    \captionsetup{justification=justified, singlelinecheck=false}
    \caption{Comparison with other image tokenizers. $^\text{oim.}$ indicates trained on OpenImages~\cite{openimage}. $\mathcal{Q}_{c}$/$\mathcal{Q}_{P}$ denotes the codebook usage in contextual and patch-level quantizers, respectively.}
    \vspace{-0.4cm}
	\label{tbl:tok_comparison2}
	\centering
    \scalebox{0.91}{
    \tablestyle{2.pt}{1.2}
	\begin{tabular}{l|c|ccc|cc|cc}
    \Xhline{0.8pt}
    \multirow{2}{*}{Method} & \multirow{2}{*}{$\textit{f}$} & \multicolumn{3}{c|}{Tokenizer Setup} & \multicolumn{2}{c|}{Image Recon.} & \multicolumn{2}{c}{Usage (\%)$\uparrow$} \\
     & ~ & Size & Dim. & \#Tok. & rFID$\downarrow$ & rIS$\uparrow$  & $\mathcal{Q}_{C}$ & $\mathcal{Q}_{P}$ \\
    \hline
    TiTok~\cite{titok} & ~--~ & 8192 & 64 & 256 & 1.05 & 191.5 
    & 100 & -- \\
    ImageFolder~\cite{imagefolder} & -- & 32768 & 32 & 286 & \textbf{0.69} & 201.5 
    & 100 & -- \\
    \hline
    $\text{VQGAN}^{\text{oim.}}$~\cite{vqgan} & \multirow{4}{*}{8} & 256 & 4 & \multirow{4}{*}{1024} & 1.44 & -- & -- & -- \\
    VQGAN~\cite{vqgan} & ~ & 8192 & 256 & ~ & 1.49 & -- & -- & -- \\
    ViT-VQGAN~\cite{vit-vqgan} & ~ & 8192 & 32 & ~ & 1.28 & 192.3 & -- & 95.0  \\
    $\text{VQGAN}^{\text{oim.}}$~\cite{vqgan} & ~ & 16384 & 4 & ~ & 1.19 & -- &  -- & --\\
    \hline
    VQGAN~\cite{vqgan} & \multirow{3}{*}{16} & \multirow{2}{*}{1024} & \multirow{2}{*}{256} & \multirow{2}{*}{256} & 7.94 & -- & --  & -- \\
    MaskGiT~\cite{maskgit} & ~ & ~ & ~ & ~ & 2.28 & -- & -- & -- \\
    VAR~\cite{var} &  & 4096 & 32 & 680 & {0.92} & 196.0  & -- & 100 \\
    \hline
    RQ-VAE~\cite{residual-vq} & 32 & 16384 & 256 & 1024 & 1.83  & -- & -- & -- \\
    \hline
    VQGAN~\cite{vqgan} & \multirow{3}{*}{16} & \multirow{3}{*}{16384} & 256 & 256 & 4.98 & --  & --  & --\\
    VQGAN~\cite{llamagen} & ~ & ~ & \multirow{2}{*}{8} & 441 & 1.21 & 189.1  & -- & 99.2 \\
    VQGAN~\cite{llamagen} & ~ & ~ &  & 576 & {0.95} & 197.3  & -- & 99.7 \\
    \hline
    \textbf{VFMTok~(\textit{Ours})}  & \multirow{1}{*}{--} & \multirow{1}{*}{16384} & \multirow{1}{*}{12} & \multirow{1}{*}{256} & {0.89} & \textbf{215.4}  & 100 & -- \\
    \Xhline{0.8pt}
	\end{tabular}}
    \vspace{-0.6cm}
\end{table}

%% file: tables/continuous_tok_eval.tex
\vspace{-.3cm}
\begin{table}[htbp]
    \captionsetup{justification=raggedleft, singlelinecheck=false} 
    \caption[Performance comparison with other continue-valued image tokenizers]{Comparison with other continue-valued image tokenizers on ImageNet validation set. L.P. indicates linear probing experiment.}
    \vspace{-0.3cm}
	\centering
    \scalebox{0.91}{
    \tablestyle{2.pt}{1.1}
	\begin{tabular}{l|c|cc|cccc|c}
    \Xhline{0.8pt}
    \multirow{2}{*}{Method} &  & \multicolumn{2}{c|}{Tok Setup} & \multicolumn{4}{c|}{Image Recon.} & L.P.\\
     & $\textit{f}$ & \#Tok.& Dim.  & rFID$\downarrow$ & rIS$\uparrow$ & PSNR$\uparrow$ & SSIM$\uparrow$ & (\%)  \\
    \hline
    SD-VAE~\cite{sd2.0} & 16 & \multirow{6}{*}{256} & 16 & 0.58 & -- & {25.7} & {0.72} & 20.4 \\ 
    VA-VAE~\cite{va-vae} & 16 & ~ & 32 & {0.28} & -- & {28.0} & 0.79 & 31.9 \\
    FAE~\cite{fae} & \multirow{4}{*}{14} & ~ & 64 & 0.66 & -- & -- & -- & 86.2  \\
    RAE~\cite{rae} & ~ & ~ & 768 & 0.57 & {226.8} & 18.8 & 0.61 & 86.7 \\
    VFMVAE~\cite{vfm-vae} & ~ & ~ & 32 & 0.52 & 214.1 & 23.0 & 0.59 & 43.2\\
    AlignTok~\cite{aligntok} & ~ & ~ & 32 & 0.26 & -- & 25.8 & -- & 35.1\\
    \hline
    VMFAE(DINOv2) & \multirow{2}{*}{16} & \multirow{2}{*}{256} & \multirow{2}{*}{32} & {0.29} & {225.6} & {25.5} & {0.82} & 80.3 \\
    VFMAE(SigLIP2) & ~ & ~ & ~ & 0.29 & 225.4 & 24.6 & 0.79 & 82.1 \\
    \Xhline{0.8pt}
	\end{tabular}}
    \vspace{-.2cm}
	\label{tbl:vfmae_img_recon}
\end{table}

%% file: tables/whole_discrete_gen.tex
\begin{table*}[ht]
  \caption{Class-conditional image generation quality estimated on ImageNet~\cite{imagenet} validation benchmark. $^{\dagger}$ indicates it is re-implemented by us, and `-re' indicates using rejection sampling.}
  \vspace{-.4cm}
  \label{tbl:distict-gen}
 \centering
    \tablestyle{1.1pt}{1.2}
  \begin{tabular}{l|l|cccccccc|ccccc}
  \Xhline{0.8pt}
 \multirow{2}{*}{Type} & \multirow{2}{*}{Method} & \multirow{2}{*}{\#Epoch} & \multirow{2}{*}{{\#Para.}} & \multirow{2}{*}{{\#Tok.}} & \multicolumn{5}{c|}{{Generation w/ CFG}} & \multicolumn{5}{c}{{Generation w/o CFG}} \\
 & &  & & & {gFID} & {sFID} & {gIS} & {Pre.} & {Rec.} & {gFID} & {sFID} & {gIS} & {Pre.} & {Rec.}  \\
  \hline
\multirow{5}{*}{Diff.} & MaskDiT~\cite{maskDiT}  & 1600 & \multirow{5}{*}{675M} & \multirow{5}{*}{256} & 2.28 & 5.67 & 276.6 & 0.80 & 0.61 & 5.69 & 10.34 & 177.9 & 0.74 & 0.60 \\
~ &  DiT~\cite{peebles2023scalable}  & 1600 & ~ &   & 2.27 & 4.60 & 278.2 & 0.83 & 0.57 & 9.62 & 6.85 & 121.5 & 0.67 & 0.67 \\
~ &  SiT-XL/2~\cite{sit} & 1600 & ~ &  &  2.06 & {4.50} & 270.3 & 0.82 & 0.59 & 8.61 & 6.32 & 131.7 & 0.68 & 0.67 \\
~ &  SiT-XL/2+REPA~\cite{repa} & 1600 & ~ & & 1.80 & 4.50 & 284.0 & 0.81 & 0.61 & 5.90 & -- & -- & -- & --  \\
~ &  FasterDIT~\cite{fasterdit} & 400 & ~ & & {2.03} & 4.63 & 264.0 & 0.81 & 0.60 & 7.91 & 5.45 & 131.3 & 0.67 & {0.69}  \\
  \hline
\multirow{2}{*}{Mask.} & MaskGiT~\cite{maskgit} & \multirow{2}{*}{555} & \multirow{2}{*}{227M} & \multirow{2}{*}{256} & -- & -- & -- & -- & -- & 6.18 & -- & 182.1 &  0.80 & 0.51  \\
 & MaskGiT-re  &  &  &  & 4.02 & -- & {355.6} &  -- & -- & -- & -- & -- & -- & -- \\
  \hline
\multirow{32}{*}{AR.}&  VAR~\cite{var} & 350 & 310M & 680 & 3.30 & -- & 274.4 & 0.84 & 0.51 & -- & -- & -- & -- & -- \\
  \cline{2-15}
 & TiTok-B$^{\dagger}$~\cite{titok} & \multirow{2}{*}{300} & 111M & \multirow{2}{*}{256} & 6.76 & 7.82 & 175.3 & {0.85} & 0.43 & 19.6 & 7.54 & 57.9 & 0.64 & 0.60 \\
 & TiTok-L$^{\dagger}$~\cite{titok}  &  & 343M &  & 4.03 & 6.93 & 219.5 & 0.84 & 0.52 & 11.4  & 7.14 & 88.8 & 0.68 & 0.64  \\
  \cline{2-15}
 & LlamaGen-B~\cite{llamagen} & \multirow{5}{*}{300} & 111M & \multirow{5}{*}{576} & 6.09 & 7.24 & 182.5 & {0.85} & 0.42 & 32.2 & 11.84 & 39.9 & 0.57 & 0.61\\
&  LlamaGen-L~\cite{llamagen}  &  & 343M &  & 3.07 & 6.09 & 256.1 & 0.83 & 0.52 & 19.1 & 8.67 & 64.3 & 0.61 &  0.67 \\  
 & LlamaGen-XL~\cite{llamagen}  &  & 775M &  & 2.63 & 5.59 & 244.1 & 0.81 & 0.58 & 15.5 & 7.04 & 79.2 & 0.62 & 0.69 \\  
 & LlamaGen-XXL~\cite{llamagen}  &  & 1.4B &  & 2.34 & 6.00 & 253.9 & 0.81 & 0.60 & 14.6 & 8.69 & 86.3 & 0.63 & 0.68 \\
 & LlamaGen-3B~\cite{llamagen}  &  & 3.1B &  & 2.19 & 5.97 & 263.3 & 0.82 & 0.58 & 9.38 & 8.24 & 112.9 & 0.69 & 0.67 \\
 \cline{2-15}
 & VFMTok-B(DINOv2) & \multirow{3}{*}{300} &  111M &  \multirow{6}{*}{256}  &  3.43 &  5.88 &  252.2 &  \textbf{0.85} &  0.53 &  3.09 &  5.67 &  173.6 &   0.80 &  0.58 \\
~ &  VFMTok-L(DINOv2) &  ~ &  343M &   &  2.75 &  5.58 &  278.8 &  0.84 &  0.57 &  2.11 &  5.46 &  230.1 &  {0.82} &  0.60 \\ 
~ & VFMTok-XL(DINOv2) &  & 775M & ~ & 2.41 & 5.53 & 276.8 & 0.83 & 0.59 & 2.10 & 5.54 & 258.1 & 0.82 & 0.60 \\
~&  VFMTok-XXL(DINOv2) &  \multirow{3}{*}{200} &  1.4B &   &  2.19 &  5.53 &  278.0 &  0.83 &  0.60 &  {1.95} &  5.65 &  259.3 &  {0.82} &  {0.62} \\
~ & VFMTok-2B(DINOv2) & & 2.2B & & 2.09 & 6.25 & 274.5 & 0.81 & 0.61 & 2.08 & 5.49 & 263.9 & 0.82 &  0.61 \\
~ &  VFMTok-3B(DINOv2) & ~ &  3.1B & ~ &  {2.07} &  6.23 &  280.4 &  0.81 &  \textbf{0.62} &  2.04 &  5.43 &  267.6 &  {0.82} &  0.61  \\
    \cline{2-15}
~& VFMTok-L(SigLIP) &300 & 343M & \multirow{3}{*}{256} & 2.61 & 5.54 & 272.1 & 0.84 & 0.56 & 2.61 & 5.54 & 272.1 & 0.84 & 0.56 \\
~& VFMTok-XXL(SigLIP) & \multirow{2}{*}{200} & 1.4B & & 2.09 & 5.75 &  272.6 & 0.82 & 0.60 & 2.09 & 5.75 & 272.6 & 0.82 & 0.60 \\
~& VFMTok-2B(SigLIP) & ~ & 2.2B & & 2.05 & 5.77 & 271.4 & 0.82 & 0.61 & 1.92 & 5.78 & 260.5 & 0.82 & 0.61 \\
\cline{2-15}
& VFMTok-L(SigLIP2) & 300 & 343M & \multirow{3}{*}{256} & 2.69 & \textbf{5.31} & 273.4 & 0.84 &  0.56 & 2.11 & {5.39} & 225.6 & 0.81 & 0.60 \\ 
& VFMTok-XXL(SigLIP2) & \multirow{2}{*}{200} & 1.4B & & 2.16 & 5.45 & 272.0 & 0.83 & 0.60 & 1.98 & 5.53 & 265.3 & {0.82} & 0.62 \\
& VFMTok-2B(SigLIP2) & ~ & 2.2B & & 2.17 & 5.43 & 281.4& 0.83 & 0.60 & 1.98 & 5.41 & \textbf{269.7} & {0.82} & 0.62 \\
 \cline{2-15}
& RAR-L~\cite{rar} & \multirow{3}{*}{400}  & 461M & \multirow{3}{*}{256} & 1.70 & -- & 299.5 & 0.82 & 0.58 & 6.72 & 5.56 & 129.2 & 0.74 & 0.61\\
  & RAR-XL~\cite{rar} &  & 955M &  & 1.50 & -- & 306.9 & 0.80 & 0.62 & 4.62 & 5.27 & 158.3 & 0.77 & 0.62\\
  & RAR-XXL~\cite{rar} &  & 1.5B &  & 1.48 & -- & {326.0} & 0.80 & 0.63 & 3.85 & 5.18 & 176.2 & 0.78 & 0.61\\
  \cline{2-15}
   & RAR-L(DINOv2) &   &  461M &   &  1.44 &  6.03 &  {312.8} &  0.78 &  0.66 &  2.02 &  5.51 &  210.4 &  0.79 &  0.63\\
   &  RAR-XL(DINOv2) &   &  955M &    &  1.38 &  5.86 &  310.2 &  0.78 &  0.65 &  1.74 &  \textbf{5.33} &  233.0 &  0.80 &  0.63\\
   &  RAR-XL(DINOv2) &  \multirow{-3}{*}{400}  &  1.5B &  \multirow{-3}{*}{256}  &  \textbf{1.36} &  5.85 &  301.3 &  0.78 &  {0.66} &  \textbf{1.65} &  5.55 &  253.7 &  0.80 &  0.63\\
   \cline{2-15}
~ & RAR-L(SigLIP) & \multirow{3}{*}{400} & 461M & \multirow{3}{*}{256} & 1.46 & 6.18 & {312.9}  & 0.78 & 0.64 & 2.26  & 5.78 & 204.3 & 0.79 & 0.62 \\
~ & RAR-XL(SigLIP) &  & 955M & & 1.40 & 6.39 & 311.7 &  0.78 & 0.65 & 1.87 & 5.77  & 226.7 & 0.79 &0.63 \\
~ & RAR-XXL(SigLIP) &  & 1.5B & & {1.38} & 6.18 & 298.4 & 0.78 & 0.65 &  1.68 & 5.64 & 239.5 & 0.79 & 0.63 \\
\cline{2-15}
~ & RAR-XL(SigLIP2) & \multirow{3}{*}{400} & 461M & \multirow{3}{*}{256} & 1.50 &  6.37 & 292.7 & 0.77 & 0.66 & 2.05 &  5.50 & 217.7 & 0.79 & 0.62  \\
~ & RAR-XL(SigLIP2) &  & 955M & & 1.46 & 6.02 & 288.4  & 0.78 & 0.65 & 1.72 &  \textbf{5.41} &  244.8 & 0.80 & 0.63 \\
~ & RAR-XXL(SigLIP2) &  & 955M & & 1.43 & 6.07 & 291.0 &  0.79  &  0.65 &  {1.65} &  5.44 & \textbf{266.5} & 0.81 & 0.63 \\
  \Xhline{0.8pt}
  \end{tabular}
  \vspace{-0.6cm}
\end{table*}

%% file: tables/continuous_gen_eval.tex
\begin{table*}[thbp]
\caption[Class-conditional image generation evaluation in continue-space]{Continue-valued class-conditional image generation on ImageNet~\cite{imagenet} validation benchmark. ${~}^{\dagger}$ indicated re-implemented by ours.} 
\vspace{-.4cm}
    \centering
    \tablestyle{1.1pt}{1.2}
    \scalebox{1.0}{
	\begin{tabular}{l|l|cccccccc|ccccc}
    \Xhline{0.8pt}
\multirow{2}{*}{Type} & \multirow{2}{*}{{Method}} & \multirow{2}{*}{\#Epoch} & \multirow{2}{*}{{\#Para.}} & \multirow{2}{*}{{\#Tok.}} & \multicolumn{5}{c|}{{Generation w/ CFG}} & \multicolumn{5}{c}{{Generation w/o CFG}} \\
 & &  & & & {gFID} & {sFID} & {gIS} & {Pre.} & {Rec.} & {gFID} & {sFID} & {gIS} & {Pre.} & {Rec.}  \\
  \hline
\multirow{28}{*}{Diff.} & MaskDiT~\cite{maskDiT}  & 1600 & \multirow{6}{*}{675M} & \multirow{6}{*}{256} & 2.28 & 5.67 & 276.6 & 0.80 & 0.61 & 5.69 & 10.34 & 177.9 & 0.74 & 0.60 \\
~&  DiT~\cite{peebles2023scalable}  & 1600 & ~ &   & 2.27 & 4.60 & 278.2 & 0.83 & 0.57 & 9.62 & 6.85 & 121.5 & 0.67 & 0.67 \\
~ &  SiT-XL/2~\cite{sit} & 1600 & ~ &  &  2.06 & {4.50} & 270.3 & 0.82 & 0.59 & 8.61 & 6.32 & 131.7 & 0.68 & 0.67 \\
~ & REPA~\cite{repa} & 1600 & ~ & & 1.80 & 4.50 & 284.0 & 0.81 & 0.61 & 5.90 & -- & -- & -- & --  \\
~&  FasterDIT~\cite{fasterdit} & 400 & ~ & & {2.03} & 4.63 & 264.0 & 0.81 & 0.60 & 7.91 & 5.45 & 131.3 & 0.67 & {0.69}  \\
~ & ReDi~\cite{redi} & 800 &  &  &  1.61 & 4.66 & 295.1 & 0.78 & 0.64 & -- & -- & -- & -- & --  \\
\cline{2-15}
~ & VA-VAE~\cite{va-vae} & 80 & \multirow{2}{*}{675M} & \multirow{2}{*}{256} &  -- & -- & -- & -- & -- & 4.29 & -- & -- & -- & --  \\ 
~ & VA-VAE~\cite{va-vae} & 800 & ~ & ~ & 1.35 & -- & 295.3 & 0.79 & 0.65 & 2.17 & -- & 205.6 & 0.77 & 0.65 \\ 
\cline{2-15}
~ & REPA-E~\cite{repae} & 80 & \multirow{2}{*}{675M} & \multirow{2}{*}{256} & -- & -- & -- & -- & -- & 7.90 & -- & 122.6 & 0.70 & 0.65 \\
~ & REPA-E~\cite{repae} & 800 & ~ & ~ & 1.29 & -- & {306.3} & 0.79 & 0.64 & 5.78 & -- & 158.3 & 0.70 & 0.68  \\
\cline{2-15}
~ & DDT~\cite{ddt} & 80 & \multirow{2}{*}{675M} & \multirow{2}{*}{256} & 1.52 & -- & 263.7 & 0.78 & 0.63 & 6.62 & -- & 135.2 & 0.69 & 0.67 \\
~ & DDT~\cite{ddt} & 800 &  & ~ &  1.26 & -- & {310.6} & 0.79 & 0.65 &  6.27 & -- & 154.7 & 0.68 & 0.69  \\
\cline{2-15}
~ & FAE~\cite{fae} & 80 & \multirow{2}{*}{675M} & \multirow{2}{*}{256}  &  1.70 & -- & 243.8 & 0.82 & 0.61 & 2.08 & -- & 207.6 & 0.82 & 0.59 \\
~ & FAE~\cite{fae} & 800 & ~ & ~  &  1.29 & -- & 268.0 & 0.80 & 0.64 & {1.48} & -- & 239.8 & 0.81 & 0.63 \\
\cline{2-15}
~ & VFMVAE~\cite{vfm-vae} & 64 & \multirow{3}{*}{685M} & \multirow{3}{*}{256} &  2.03 & 5.23 & 261.7 & 0.83 &0.58  & 2.42 & -- & 215.2 & -- & -- \\ 
~ & VFMVAE~\cite{vfm-vae} & 80 & ~ & ~ & -- & -- & -- & -- & -- & 2.22 & 5.03 & 218.8 & 0.83 & 0.58 \\
~ & VFMVAE~\cite{vfm-vae} & 640 & ~ & ~ & 1.31 & 4.63 & 300.2 & 0.78 & 0.66 & 1.62 & 4.55 & 241.6 & 0.81 & 0.62 \\
\cline{2-15}
~ & AlignTok~\cite{aligntok} & 64 & \multirow{2}{*}{675M} & \multirow{2}{*}{256} & 1.90 & -- & 260.9 & 0.81 & 0.61 & 3.71 & -- & 148.9 & 0.77 & 0.62\\
~ & AlignTok~\cite{aligntok} & 800 & ~ & ~ & 1.37 & -- & 293.6 & 0.79 & 0.65 & 2.04 & -- & 206.2 & 0.76 & 0.67 \\
\cline{2-15}
~ & RAE~\cite{rae} & 80 & \multirow{5}{*}{839M} & \multirow{5}{*}{256} & 2.16 & -- & 214.8 & 0.82 & 0.59 & -- &  -- & --  & -- & --  \\ 
~ & RAE~\cite{rae} & 1400 & ~ & ~ & {1.13} & -- & 262.6 & 0.78 & 0.67 & 1.51 & -- & 242.9 & 0.79 & 0.63  \\
~ & $\text{RAE}^{\dagger}$~\cite{rae} & 80 & ~ & ~ & 1.76 & 4.58 & 221.7 & 0.80 & 0.61 & 2.28 & 5.26 & 203.0 & 0.81 & 0.60 \\
~ & $\text{RAE}^{\dagger}$~\cite{rae} & 800 & ~ & ~ & 1.27 & 4.70 & 248.5 & 0.76 & 0.67 & 1.60 & 5.26 & 232.3 & 0.78 & 0.65 \\
~ & $\text{RAE}^{\dagger}$~\cite{rae} & 1400 & ~ & ~ & 1.19 & 4.65 & 254.6 & 0.77 & 0.67 & {1.53} & 5.15 & 235.9 & 0.78 & 0.64 \\
\cline{2-15}
~ & VFMAE(DINOv2) & 80 & \multirow{4}{*}{839M} & \multirow{4}{*}{256} & 1.68 & 6.10 & 217.5 & 0.74 & 0.68 & 3.29 & 6.17 & 176.5 & 0.75 & 0.64  \\
~ & VFMAE(DINOv2) & 800 & ~ & ~ & {1.25} & 5.84 & {294.0} & 0.77 & 0.69 & 1.65 & 5.78 & {240.9} & 0.76 & 0.67  \\
~ & VFMAE(SigLIP2) & 80 & ~ & ~ & 1.40 & 5.65 & 224.2 & 0.76 & 0.67 & 3.34 & 5.93 & 172.1 & 0.75 & 0.65  \\
~ & VFMAE(SigLIP2) & 800 & ~ & ~ & 1.29 & 5.93 & 280.9 & 0.76 & 0.68 & 1.70 & 5.75 & 235.4 & 0.76 & 0.68  \\
  \Xhline{0.8pt}
	\end{tabular}}
\vspace{-0.4cm}
\label{tbl:continue_gen2}
\end{table*}

%% file: tables/core_comp_analysis.tex
 \begin{table*}
 \begin{minipage}{0.5\textwidth}
    \caption{Ablation study on {\ours}'s core components.}
	\label{tbl:ablation_queries}
    \vspace{-.23cm}
	\centering
    \scalebox{0.96}{
    \vspace{-.5em}
    \tablestyle{1.8pt}{1.05}
	\begin{tabular}{l|ccc|c|cc|c}
		\Xhline{0.8pt}
    \multirow{2}{*}{Setup} & \multicolumn{3}{c|}{Image Recon.} & \multicolumn{1}{c|}{Usage} & \multicolumn{2}{c|}{AR Gen.} & L.P. \\
    & \#Tok. & rFID$\downarrow$ & rIS$\uparrow$ & $\mathcal{Q}_{C}\uparrow$ & gFID$\downarrow$ & gIS$\uparrow$ & (\%) \\
    \hline
    VQGAN & \multirow{2}{*}{576} &  0.95 & 197.3  & {99.7\%} & 3.71 & 228.3 & 23.1 \\
    + {Frozen VFM} &  & 0.99 & 206.3 &  100\% & 3.69 & 267.5 & 56.4 \\
    \hline
    + {Region Adapt.} & \multirow{3}{*}{256} & 1.20 & 199.0 & \multirow{3}{*}{100\%} & 3.98 & 241.6 & 41.5 \\
    + {Multi-level Feat.} &  & 0.92 & 199.5 & & {3.71} & {251.1} & {22.7} \\
    + {Reconstruct Feat.} &  & \textbf{0.89} & \textbf{215.4} &  & \textbf{3.42} & \textbf{277.3} & \textbf{69.4}\\
    \hline
    - {Frozen VFM} & 256 & 0.95 & 196.3 & 100\% & 3.73 & 248.7 & 59.1 \\
    \Xhline{0.8pt}
	\end{tabular}}
    \end{minipage}
    \hfill
    \begin{minipage}{0.5\textwidth}
    \caption{Impact study on learnable components. CA denotes Cross-attention while DA indicates Deformable-attention. $\text{M}_\text{share}$ represents shared mask token.}
	\label{tbl:part_annalysis}
    \vspace{-.2cm}
	\centering
    \scalebox{1.0}{
    \tablestyle{1.5pt}{1.2}
	\begin{tabular}{cccc|c|cc|cc|c}
		\Xhline{0.8pt}
           \multirow{2}{*}{$\text{M}_\text{shared}$}  & \multicolumn{2}{c}{Attention} & \multirow{2}{*}{$\text{Tok}_\text{reg}$} & \multicolumn{3}{c|}{$\textit{Image Recon.}$} & \multicolumn{2}{c|}{$\textit{AR gen.}$} & {L.P.} \\
            \cline{2-3} \cline{5-9}
            & CA & DA & & \#Toks & rFID$\downarrow$ & rIS$\uparrow$ & gFID$\downarrow$ & gIS$\uparrow$ & (\%)  \\
            \hline
  ${\checkmark}$  & ${\checkmark}$ & & ${\checkmark}$ & \multirow{4}{*}{256} & 1.00 & 211.5 & 3.89 & 271.5 & 34.1 \\
 \checkmark   &  & \checkmark & &  & 0.91 & 215.8 & 3.45 & 276.5  & 68.3 \\
     & & \checkmark & \checkmark &  & 0.89 & {216.1} & 3.50 & 276.0 & 64.7 \\
   \checkmark & & \checkmark & \checkmark &  & {0.89} & {215.4} & {3.42} & {277.3} & {69.4} \\
		\Xhline{0.8pt}
	\end{tabular}}
    \end{minipage}
\vspace{-.5cm}
\end{table*}

%% file: tables/feat_analysis_vfm_category.tex
\begin{table*}[htbp]
\centering
\begin{minipage}{0.36\textwidth}
\centering
\caption{Performance of each single-level feature. $F_i$ represents the indexed feature level.}
\vspace{-0.35cm}
\label{tbl:each-single-level}
\vspace{3pt}
\scalebox{0.99}{
\tablestyle{1.5pt}{1.1}
\begin{tabular}{l|c|cc|c|cc}
\Xhline{0.8pt}
\multirow{2}{*}{$F_i$}  & \multicolumn{4}{c|}{$\textit{Image recon.}$} & \multicolumn{2}{c}{$\textit{AR gen.}$} \\
\cline{2-5}\cline{6-7}
&  \#Toks & rFID$\downarrow$ & rIS$\uparrow$ & $\mathcal{Q}_{C}$ & gFID$\downarrow$ & gIS$\uparrow$  \\
\hline
$F_1$ & \multirow{4}{*}{256} & 1.04 & 186.4 & \multirow{4}{*}{100.0\%} & 3.84 & 257.9 \\
$F_2$ &   & 0.95 & 200.6 &  & 3.79	& 272.7 \\
$F_3$ &   & 1.03 & 208.5 & & 3.69 & 274.9 \\
$F_4$ &   &1.23  & 214.8 &   & 3.64 & 277.7  \\
\Xhline{0.8pt}
\end{tabular}}
\end{minipage}
\hfill
\begin{minipage}{0.36\textwidth}
\centering
\caption{Performance of cumulatively added features. $F_i$ represents the indexed feature level.}
\vspace{-0.35cm}
\label{tbl:cumulative-features}
\vspace{3pt}
\scalebox{0.99}{
\tablestyle{1.5pt}{1.1}
\begin{tabular}{l|c|cc|c|cc}
\Xhline{0.8pt}
\multirow{2}{*}{$F_i$}  & \multicolumn{4}{c|}{$\textit{Image recon.}$} & \multicolumn{2}{c}{$\textit{AR gen.}$} \\
\cline{2-5}\cline{6-7}
&  \#Toks & rFID$\downarrow$ & rIS$\uparrow$ & $\mathcal{Q}_{C}$ & gFID$\downarrow$ & gIS$\uparrow$  \\
\hline
$F_1$ & \multirow{4}{*}{256} & 1.04 & 186.4 & \multirow{4}{*}{100.0\%} & 3.84 & 257.9 \\
$+F_2$ &   & 0.94 & 205.0 &  & 3.69	& 274.4 \\
$+F_3$ &   & 0.94 & 210.8 & & 3.27 & 272.5 \\
$+F_4$ &   & 0.89  & 215.4 &  & 3.09 & 274.2  \\
\Xhline{0.8pt}
\end{tabular}}
\hfill
\end{minipage}
\begin{minipage}{0.27\textwidth}
\caption{The category of VFMs based on their learning objectives.}
\vspace{-0.35cm}
	\label{tbl:vfms_objective_grouping}
	\centering
    \scalebox{0.96}{
    \tablestyle{1.1pt}{1.05}
	\begin{tabular}{l|ccc}
		\Xhline{0.8pt}
        VFM & P-MIM & L-MIM & C.L. \\
        \hline
        CLP~\cite{clip} & \ding{55} & \ding{55} & \checkmark \\
        DINO~\cite{dino} & \ding{55} & \ding{55} & \checkmark  \\
        SigLIP~\cite{siglip} &  \ding{55} & \ding{55} & \checkmark \\
        MAE~\cite{mae} & \checkmark & \ding{55} & \ding{55} \\
        iBOT~\cite{ibot} & \ding{55} & \checkmark & \ding{55} \\
        SigLIP2~\cite{siglip2} & \ding{55} & \checkmark & \checkmark \\
        DINOv2~\cite{dinov2reg} & \ding{55} & \checkmark & \checkmark \\
		\Xhline{0.8pt}
	\end{tabular}}
\end{minipage}
\vspace{-0.4cm}
\end{table*}

%% file: tables/vfm_objectives.tex
\begin{table*}[ht]
	\caption{Image reconstruction and generation with VFM of a distinct self-supervised learning objective.}
    \vspace{-0.4cm}
	\label{tbl:mim_gen}
	\centering
    \scalebox{0.95}{
    \tablestyle{1.85pt}{1.05}
	\begin{tabular}{l|l|ccc|ccccc|ccccc|c}
    \Xhline{0.8pt}
    \multirow{2}{*}{Type} & \multirow{2}{*}{Tokenizer} & \multirow{2}{*}{Pixel-MIM} & \multirow{2}{*}{Latent-MIM} & \multirow{2}{*}{C.L.} & \multicolumn{5}{c|}{$\textit{Image recon.}$} & \multicolumn{5}{c|}{\textit{Image gen.}} & \multirow{2}{*}{L.P.(\%)$\uparrow$} \\ 
    ~ &  ~ & ~ & ~ & ~ & \#tok. & rFID$\downarrow$ & rIS$\uparrow$ & PSNR$\uparrow$ & SSIM$\uparrow$ & gFID $\downarrow$ & sFID$\downarrow$ & gIS$\uparrow$ & Pre.$\uparrow$ & Rec.$\uparrow$ & ~ \\ 
    \hline
\multirow{6}{*}{AR.} & VQGAN(CLIP~\cite{clip}) & \ding{55} & \ding{55} & \checkmark & \multirow{6}{*}{576} & 1.47 & 182.0 & 17.8 & 0.58 & 3.45 & 5.59 &221.2 & 0.86 & 0.50  &59.5 \\
~ & VQGAN(SigLIP~\cite{siglip}) & \ding{55} & \ding{55} & \checkmark & & 1.26 & 190.8 & 19.4 & 0.60 & 3.50 & 5.78 & 246.1 & 0.87 & 0.52 &  60.3 \\
~ & VGQGAN(MAE~\cite{mae}) & \checkmark & \ding{55} & \ding{55} & & 0.88 & 190.4 & 21.4 & 0.72 & 3.33 & 5.83 & 250.7  & 0.87 & 0.52	& 42.1 \\
~ & VQGAN(iBOT~\cite{ibot}) & \ding{55} & \checkmark & \ding{55} & & 1.10 & 183.4& 19.8 & 0.66 & 3.39 & 5.53 & 247.2 & 0.87 & 0.51 & 48.5 \\
~ & VQGAN(SigLIP2~\cite{siglip2}) & \ding{55} & \checkmark & \checkmark & & 0.96 &  198.4 & 18.7 & 0.62 & 3.39 & 5.58 & 267.8 &0.87 & 0.51 & 55.5 \\
~ & VQGAN(DINOv2~\cite{dinov2reg}) & \ding{55} & \checkmark & \checkmark &  & 0.99 & 206.3& 18.4 & 0.59 & 3.34 & 5.52 & 268.6 &0.87 & 0.51& 56.4 \\
    \hline
\multirow{6}{*}{Diff.} & AE(CLIP~\cite{clip}) & \ding{55} & \ding{55} & \checkmark & \multirow{6}{*}{256} & 1.13 & 199.2 & 18.1 & 0.57 & 3.14 &  5.92 & 210.3 & 0.85 & 0.53 & 62.5 \\
~ & AE(SigLIP~\cite{siglip}) & \ding{55} & \ding{55} & \checkmark & ~ & 1.04 & 211.8 & 17.5 & 0.55 & 2.78 &  5.91 &  214.0 & 0.84 & 0.55 & 63.1 \\
~ & AE(MAE~\cite{mae}) & \checkmark & \ding{55} & \ding{55} & ~ & 0.31 & 223.6 & 25.4 & 0.82 & 2.52 & 5.96 & 239.7 & 0.83 & 0.57 & 36.7 \\
~ & AE(iBOT~\cite{ibot}) & \ding{55}  & \checkmark & \ding{55} & ~ & 0.53 & 212.9 & 21.6 & 0.72 & 2.51 & 5.63 & 226.7 & 0.83 & 0.57 & 55.9 \\
~ & AE(SigLIP2~\cite{siglip2}) & \ding{55} & \checkmark & \checkmark & ~ & 0.74 & 219.6 & 19.1 & 0.62 & 2.45 & 5.87 & 223.1 & 0.82 & 0.58 & 61.2 \\
~ & AE(DINOv2~\cite{dinov2reg}) & \ding{55} & \checkmark & \checkmark & ~ & 0.79 & 228.1& 18.2 & 0.58 & 2.30 & 5.79 & 239.0 & 0.83 & 0.58 & 61.5 \\
  \Xhline{0.8pt}
	\end{tabular}}
\vspace{-.6cm}
\end{table*}

%% file: sections/5_conclusion.tex
\vspace{-0.2cm}
\section{Conclusion}

In this work, we establish that frozen pre-trained vision foundation models (VFMs)---regardless of whether they are self- or language-supervised---serve as exceptionally capable visual tokenizers for both discrete and continuous latent spaces. To unlock their full potential and address the redundancy inherent in traditional 2D feature grids, we propose \ours. By leveraging region-adaptive sampling, multi-level features integration, and a semantic-preserving feature reconstruction objective, \ours produces a compact, semantically rich latent space. Consequently, this architecture not only facilitates superior image reconstruction and synthesis but also accelerates the convergence in autoregressive (AR) models. Additionally, it enables efficient, high-fidelity image generation without relying on classifier-free guidance (CFG-free) — without requiring additional training heuristics. Notably, beyond empirical experiments, we delve into the fundamental reasons why VFM is good for image tokenizer design. Through an extensive investigation of various VFM pre-training objectives, we identify that a VFM jointly trained with Contrastive Learning (C.L.) and latent Masked Image Modeling (Latent-MIM) provides the optimal foundational backbone for image tokenizer design.

%% file: sections/6_appendix.tex
\section*{Appendix}
This section first illustrates the implementation of discrete VFMTok and continuous-valued VFMAE, as well as relevant generative models, covering their learning rates, optimization approaches, and training requirements. Subsequently, we present additional ablation studies on VFMTok and its variant VFMAE, examining the impact of various architectural designs on image reconstruction and synthesis. Additionally, we present additional visualization samples of class-to-image generation using VFMTok.

\section{VFMTok Implementation}
\noindent\textbf{Discrete VFMTok training.} VFMTok is trained on the ImageNet~\cite{imagenet} training set at $336\times336$ resolution with random crop augmentation. All models share identical training settings: a constant learning rate of ${10}^{-4}$, AdamW optimizer~\cite{adamw} (${\beta}_1 = 0.9, {\beta}_2 = 0.95$, weight decay = 0.05), a batch size of 256, and 50 training epochs. For training losses, the commitment loss weight is 0.25 and the adversarial loss weight is 0.5, with the adversarial loss activated after 20,000 iterations. Besides, VFMTok requires 1.5 days of training on 16 Nvidia H800 GPUs.

\noindent\textbf{Continuous-Valued VFMAE training.} VFMAE is also trained on the ImageNet~\cite{imagenet} training set and estimated on the validation set, but at the image resolution of $256\times256$ with random crop augmentation. All models share the same training settings: a constant learning rate of ${10}^{-4}$, AdamW optimizer~\cite{adamw} (${\beta}_1 = 0.9, {\beta}_2 = 0.95$, weight decay = 0.05), and 50 training epochs. The adversarial loss weight is set to 0.5, and it is activated after 20,000 iterations. For efficiency, the batch size for VFMAE is increased to 512. The training lasts for 1 day when training on 16 Nvidia H800 GPUs.

\noindent\textbf{AR model optimization.} The training configuration aligns with LlamaGen's~\cite{llamagen}, except our training resolution is $336\times336$ and the duration depends on model parameters. Other key settings include: a base learning rate of ${10}^{-4}$ per 256 batch size; AdamW optimizer~\cite{adamw} (${\beta}_{1} = 0.9$, $\beta_{2} = 0.95$, weight decay = 0.05, gradient clipping of 1.0); a dropout rate of 0.1 for input token embeddings, attention, and FFN modules; and a 0.1 class condition embedding dropout for classifier-free guidance. Besides, A VFMTok-L requires 19.4 hours of 50 epochs training on 8 NVIDIA H800 GPUs.

\noindent\textbf{Denoising Generative Model Optimization.} The model architecture is generally inherited from RAE's~\cite{rae} $\text{DiT}^\text{DH}-\text{XL}$ across all $256\times256$ experiments, and the model processes a token sequence of length 256. Meanwhile, we first compute the mean and variance of the latent features over 400,000 training images. Then we utilize them to normalize the latent features of VFMAE during training the diffusion model. Besides, we follow the optimization strategy in RAE, using AdamW~\cite{adamw} with a linear decay from $2.0\times10^{-4}$ to $2.0\times10^{-5}$ with a constant warming-up of 40 epochs. We also utilize an EMA weight of 0.9995 to update the EMA model. Moreover, we use gradient clipping of 1.0 for those models. Additionally, we set the batch size to 512. Other optimization hyperparameters remain identical to those of RAE~\cite{rae}. To maintain a fair comparison with most denoising models, we train all models for at most 800 epochs and only report the EMA model’s generation performance at 80 and 800 epochs. During inference, we use standard ODE sampling with the Euler sampler and 100 steps by default.
\section{Supplemental Ablation Study}

In this subsection, we conduct more ablation studies on the design of VFMTok, including the AR generation with resolution of $256\times256$, the effect of shared ViT \textit{v.s.} unshared ViT for discrete VFMTok design, and the number of tokens to represent an image for both reconstruction and synthesis.

\subsection{Discrete-Space Generation with Size of $256\times256$.}
\input{tables/discrete_gen_256x256}
In the main manuscript, given that the input resolution of the vision foundation model is $336\times336$, we adjust the resolution of reconstructed and generated images to $336\times336$ by default, thus avoiding changing the number of tokens for image representation. Following the common practices~\cite{llamagen, var}, we also train the image tokenizer and the AR generation model with the resolution of $256\times256$, respectively. We first initialize and train VFMTok tokenizer for 50 epochs, then integrate it with AR generative models. Considering computational costs, models with fewer parameters like VFMTok-B and VFMTok-L are trained for 300 epochs, while larger AR models for 50 epochs. This setting aligns with LlamaGen~\cite{llamagen} for $256\times256$ image generation. Furthermore, to ensure a fair comparison with the advanced autoregressive generation framework, RAR~\cite{rar}, we also incorporated VFMTok into RAR~\cite{rar}, maintaining the same setup during the training phase. As shown in \cref{tbl:low_resolution}, \ours not only achieves a decent reconstruction performance but also improves the generation quality compared to its counterparts~\cite{llamagen, rar} by a large margin. It is worth noting that \ours also accelerates the convergence speed during AR model training and significantly improves synthesis quality.

\subsection{Effect of Shared ViT \textit{v.s.} Unshared ViT}

In this work, VFMTok utilizes a shared ViT to generate latent features for pixel rendering and high-level VFM feature (specifically, from the last layer) reconstruction, respectively. However, it is uncertain if the sharing parameter is optimal. To this end, we experimented with another unshared ViT of the same architecture to generate the high-level VFM feature. Following the training setup, we train the tokenizer and AR model -- VFMTok-L for 50 epochs. As shown in Tab.~\ref{tbl:shared-vs-unshared}, the shared ViT ensures a better image reconstruction and synthesis quality with enriched semantics. Therefore, we utilize shared ViT in the VFMTok design by default.

\input{tables/vit_ablation}

\subsection{Effect of the Token Count to Represent an Image.}
In this work, we utilize a set of region-adaptive tokens to represent an image for autoregressive (AR) image generation. The number of these tokens is empirically fixed at 256 throughout our experiments. However, the impact of tokens' cardinality on image reconstruction and generation quality remains unexplored.
\input{tables/tok_cardinality}

To investigate this, we designed a controlled setup to isolate the impact of query quantity while maintaining architectural consistency across experiments. Specifically, we parameterize the image tokenizer with varying numbers of learnable queries. Each tokenizer variant is trained on the ImageNet~\cite{imagenet} training set for 50 epochs. Subsequently, the trained tokenizer is integrated into an AR image generation model, LlamaGen-L~\cite{llamagen}, and trained for 50 epochs. Both image reconstruction and generation quality are estimated on ImageNet~\cite{imagenet} validation set with FID and IS, respectively. 

As shown in~\cref{tbl:num_toks}, image reconstruction exhibits a positive correlation with the tokens' cardinality. As the number of tokens increases, metrics for estimating image reconstruction, namely rFID, rIS, PSNR, and SSIM, all show gradual improvement. However, this trend does not hold for generation: as the number of tokens increases to higher values, there is no significant improvement in generation quality. Therefore, to balance generation quality with computational cost, and maintain fairness with vanilla counterparts~\cite{vqgan, vqgan-lc, llamagen}, we fix the number of tokens at 256 across our experiments. Actually, it's observed that \textbf{144} visual tokens suffice for representing images for image synthesis in ImageNet~\cite{imagenet}. This finding indicates VFMTok can further eliminate redundancy within image representations, yielding more compact and more effective image compression.

\subsection{More Visual Samples.}
\begin{figure*}[htbp]
  \centering
  \includegraphics[width=0.85\textwidth]{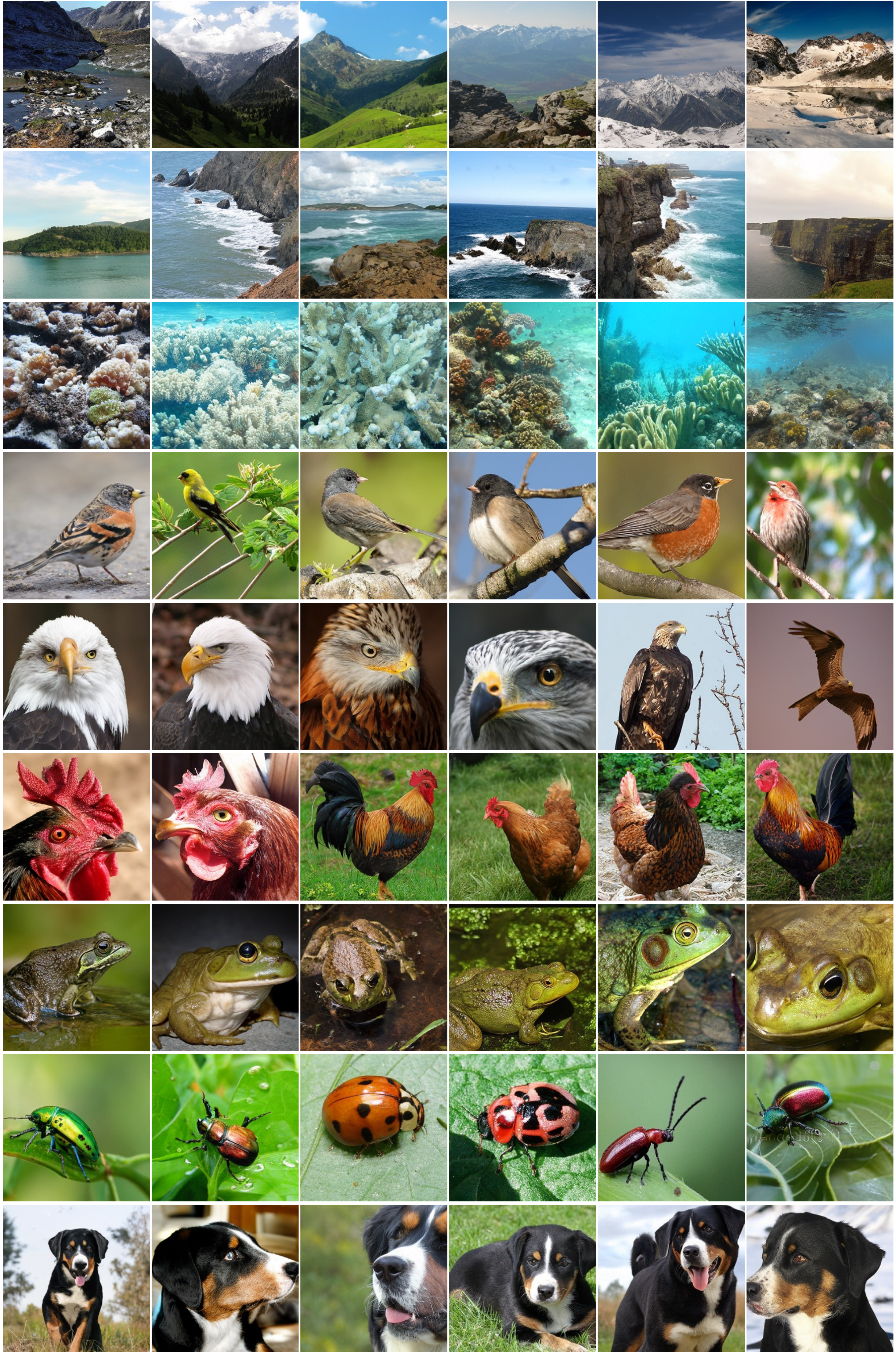}
  \vspace{-0.3cm}
  \text{Fig.3. Autoregressive class-conditional image generation with classifier-free guidance (CFG).}
  \label{fig:supp_gen_cfg}
\end{figure*}

\begin{figure*}[thbp]
  \centering
  \includegraphics[width=0.85\textwidth]{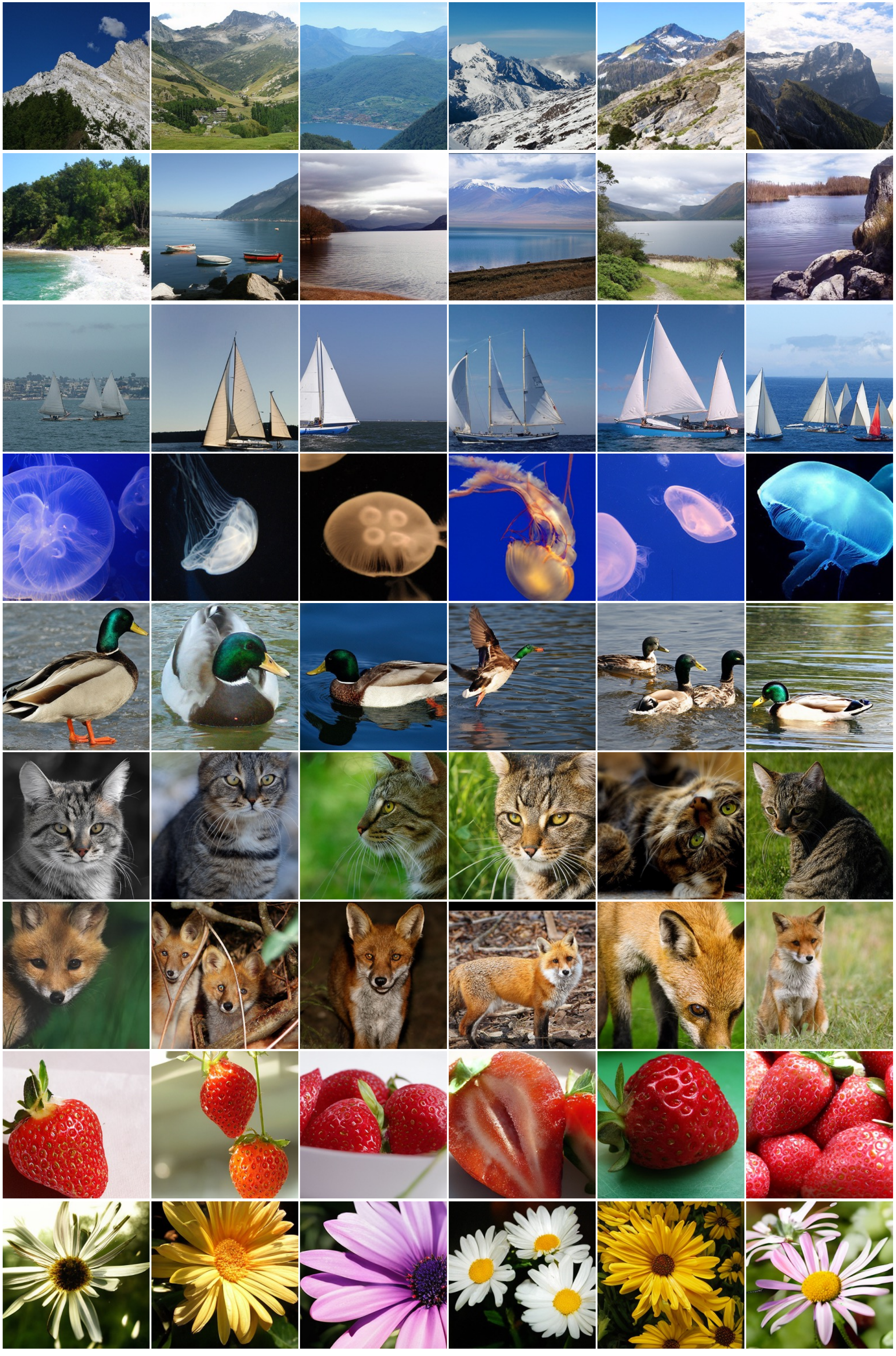}
  \vspace{-0.3cm}
  \text{Fig.4. Autoregressive class-conditional image generation without classifier-free guidance (CFG).}
  \label{fig:supp_gen_wo_cfg}
\end{figure*}

In this section, we present additional visualization samples related to autoregressive class-to-image image generation with (see Fig.~\ref{fig:supp_gen_cfg} and without classifier-free guidance (see Fig.~\ref{fig:supp_gen_wo_cfg}). For optimal clarity, please zoom in.

%% file: tables/discrete_gen_256x256.tex
\begin{table}[ht]
	\caption{\ours performs {image reconstruction} and {AR generation} with the size of $256\times256$. $\text{R}^{\star}$ is the abbreviation of RAR~\cite{rar}.}
	\label{tbl:low_resolution}
	\centering
    \scalebox{0.94}{
    \tablestyle{1.05pt}{1.1}
	\begin{tabular}{l|c|cc|c|c|c|cc}
		\Xhline{0.8pt}
            \multirow{2}{*}{Approach} & \multicolumn{3}{c|}{$\textit{Image recon.}$} &\multirow{2}{*}{$\mathcal{Q}\uparrow$} & \multicolumn{4}{c}{$\textit{AR gen.}$} \\
            \cline{2-4}\cline{6-9}
            & \#Toks & rFID$\downarrow$ & rIS$\uparrow$ & ~ & \#$\text{E}$ & Para. & gFID$\downarrow$ & gIS$\uparrow$   \\
            \hline
            LlamaGen-B  & \multirow{5}{*}{256} & \multirow{5}{*}{2.22} &\multirow{5}{*}{169.8} & \multirow{5}{*}{95.2\%} & \multirow{2}{*}{300} & 111M & 5.46 & 193.6  \\
            LlamaGen-L & &  &  &  & & 343M  & 3.81 & 248.3  \\
            \cline{7-7}
            LlamaGen-XL & &  &  &  & \multirow{3}{*}{50} & 775M  & 3.39 & 227.1  \\
            LlamaGen-XXL & &  &  &  & & 1.4B  & 3.09 & 253.6  \\
            LlamaGen-3B & & & & & & 3.1B & 3.06 & 279.7  \\
            \hline
            RAR-L~\cite{rar} & \multirow{3}{*}{256} & \multirow{3}{*}{2.12} & \multirow{3}{*}{171.4} & \multirow{3}{*}{100\%}  & \multirow{3}{*}{400} & 461M & 1.70 & 299.5  \\
            RAR-XL~\cite{rar} & & & & & & 955M & 1.50 & 306.9 \\
            RAR-XXL~\cite{rar} & & & & & & 1.5B & 1.48 & \textbf{326.0} \\
            \hline
            \ours-B  & \multirow{7}{*}{256} & \multirow{7}{*}{1.02} & \multirow{7}{*}{213.2}  &\multirow{7}{*}{100\%} &  \multirow{2}{*}{100} & 111M & {3.95} & {248.4} \\
            \ours-L & & &  &  &  & 343M  & {3.02} & {271.6} \\
            \cline{7-7}
            \ours-B & & &  &   & \multirow{2}{*}{300} & 111M  & {3.61} & {247.6} \\
            \ours-L & & &  & & & 343M  & {2.79} & {276.0} \\
            \cline{7-7}
            \ours-XL & & &   & & \multirow{3}{*}{50} & 775M  & {2.79} & {277.1} \\
            \ours-XXL & &  &  & & & 1.4B & {2.62} & {279.7} \\
            \ours-2B & & & & & & 2B & {2.64} & {284.0}\\
            \hline
            VFMTok($\text{R}^{\star}$-L) & \multirow{3}{*}{256} & \multirow{3}{*}{0.88} & \multirow{3}{*}{216.2} & \multirow{3}{*}{100\%}  & \multirow{3}{*}{400} & 461M & 1.47 & {316.2}  \\
            VFMTok($\text{R}^{\star}$-XL) & & &  & & & 955M & \underline{1.38} & 303.3 \\
            VFMTok($\text{R}^{\star}$-XXL) & & & & & & 1.5B & \textbf{1.30} & 300.0 \\
		\Xhline{0.8pt}
	\end{tabular}}
\end{table}

%% file: tables/vit_ablation.tex
\begin{table}[htbp]
\vspace{-0.4cm}
	\caption{Effect of shared ViT \textit{v.s.} unshared ViT.}
	\label{tbl:shared-vs-unshared}
	\centering
    \vspace{-0.23cm}
    \tablestyle{1.2pt}{1.1}
	\begin{tabular}{l|c|cc|c|c|cc|c}
		\Xhline{0.8pt}
           \multirow{2}{*}{Type}  & \multicolumn{3}{c|}{$\textit{Image recon.}$} &\multirow{2}{*}{$\mathcal{Q}_{C}$} & \multicolumn{3}{c|}{$\textit{AR gen.}$} & \multirow{2}{*}{L.P.} \\
            \cline{2-4}\cline{6-8}
            &  \#Toks & rFID$\downarrow$ & rIS$\uparrow$ &   & \#${E}$ & gFID$\downarrow$ & gIS$\uparrow$ &  \\
            \hline
    unshared ViT  & \multirow{2}{*}{256} & {0.91} & {214.9} & {100.0\%} & 50 & 3.52 & {277.4} & {61.1}  \\
    \cline{1-1}\cline{3-9}
    shared ViT &  & {0.89} & {215.4} & {100.0\%} & 50 & {3.42} & 277.3 & {69.4} \\
		\Xhline{0.8pt}
	\end{tabular}
    \vspace{-.5cm}
\end{table}

%% file: tables/tok_cardinality.tex
\begin{table}[htbp]
\caption[The effects of token counts on image reconstruction and generation]{The effects of tokens count to represent an image.}
\centering
    \tablestyle{1.2pt}{1.05}
    \scalebox{0.98}{
    \begin{tabular}{c|cccc|ccccc}
        \Xhline{0.8pt}
           \multirow{2}{*}{\#Toks.}  & \multicolumn{4}{c|}{$\textit{Image recon.}$}  & \multicolumn{5}{c}{$\textit{AR gen.}$} \\
            \cline{2-10}
            &   rFID$\downarrow$ & rIS$\uparrow$ & PSNR$\uparrow$ & SSIM$\uparrow$ &  gFID$\downarrow$ & gIS$\uparrow$ & sFID$\downarrow$ & Pre.$\uparrow$ & Rec.$\uparrow$  \\
            \hline
    36 & 2.61 & 175.4 & 17.0 & 0.51 & 3.93 & 222.4 & 6.07 & 0.86 & 0.48   \\
    64 & 2.09 & 188.3 & 18.0 & 0.55 & 3.59 & 250.5 & 5.89 & 0.86 & 0.51   \\
    100 & 1.44 & 204.4 & 18.9 & 0.59 & 3.54 & 270.8 & 5.60 & 0.86 & 0.52 \\
    121 & 1.36 & 202.5 & 19.2 & 0.60 & 3.59 & 267.1 & 5.55 & 0.86 & 0.53 \\
    {144} & {1.20} & {204.6} & 19.5 & 0.62 & {3.46} & {274.9} & 5.64 & 0.85 & 0.54 \\
    169 & 1.11 & 212.4 & 19.9 & 0.63 & 3.42 & 275.5 & 5.40 & 0.86 & 0.53 \\
    196 & 1.08 & 213.7 & 20.0 & 0.64 & 3.32 & 271.2 & 5.86 & 0.85 & 0.54 \\
    225 & 0.96 & 215.5 & 20.4 & 0.65 & 3.45 & 279.2 & 5.59 & 0.86 & 0.53 \\
    256 & 0.89 & 215.4 & 20.6 & 0.67 & 3.42 & 277.3 & 5.68 & 0.85 & 0.53 \\
    289 & 0.85 & 217.2 & 20.9 & 0.68 & 3.41 & 272.3 & 5.30 & 0.87 & 0.52 \\
    324 & 0.88 & 219.8 & 20.8 & 0.68 & 3.42 & 272.3 & 5.76 & 0.85 & 0.53 \\
    361 & 0.80 & 218.8 & 21.3 & 0.69 & 3.64 & 277.9 & 5.20 & 0.86 & 0.52 \\
    400 & 0.79 & 221.3 & 21.3 & 0.70 & 3.36 & 272.3 & 5.31 & 0.85 & 0.54  \\
    441 & 0.73 & 221.8 & 21.5 & 0.71 & 3.61 & 275.5 & 5.29 & 0.86 & 0.51 \\ 
    576 & 0.60 & 222.8 & 22.3 & 0.74 & 3.57 & 278.8 & 5.71 & 0.85 & 0.52 \\
        \Xhline{0.8pt}
	\end{tabular}
    }
	\label{tbl:num_toks}
\end{table}